%% file: srm-tmp.tex
\documentclass{article}

\usepackage{microtype}
\usepackage{graphicx}
\usepackage{subfig}
\usepackage{booktabs} 
\usepackage{pgfplots}
\usepgfplotslibrary{groupplots}
\usetikzlibrary{pgfplots.colormaps, arrows.meta, positioning, calc, math}
\pgfplotsset{compat=1.18}

\usepackage{hyperref}

\usepackage[preprint]{icml2026}

\hypersetup{
  pdftitle={Stream Recursion Model (SRM)},
  pdfsubject={Machine Learning; Mechanistic Interpretability},
  pdfauthor={David Chamberlain; Ramyaa Ramyaa; Asael H Sorensen; Charles
  Brock; Jennifer Minnich; Matthew J. Hoffman},
}

\usepackage{amsmath}
\usepackage{amssymb}
\usepackage{mathtools}
\usepackage{amsthm}
\usepackage{listings}

\usepackage{soul}

\usepackage[capitalize,noabbrev]{cleveref}

\theoremstyle{plain}

\theoremstyle{definition}

\theoremstyle{remark}

\usepackage[disable,textsize=tiny]{todonotes}

\icmltitlerunning{Stream Recursion Model (SRM)}

\begin{document}

\twocolumn[
  \icmltitle{Stream Recursion Model (SRM)}

  \icmlsetsymbol{equal}{*}

  \begin{icmlauthorlist}
    \icmlauthor{Asael Sorensen}{snl}
    \icmlauthor{Charles Brock}{icasa,nmt,equal}
    \icmlauthor{David Chamberlain}{icasa,nmt,equal}
    \icmlauthor{Jennifer Minnich}{icasa,equal}
    \icmlauthor{Matthew Hoffman}{snl}
    \icmlauthor{Ramyaa Ramyaa}{nmt}
  \end{icmlauthorlist}

  \icmlaffiliation{nmt}{Department of Computer Science and Engineering,
  New Mexico Institute of Mining and Technology, Socorro, NM, USA}
  \icmlaffiliation{icasa}{Institute for Complex Additive Systems
    Analysis, New Mexico Institute of Mining and Technology, Socorro,
  NM, USA}
  \icmlaffiliation{snl}{Sandia National Laboratories, Albuquerque, NM,
  USA}

  \icmlcorrespondingauthor{David Chamberlain}{david.chamberlain@student.nmt.edu}

  \icmlkeywords{Machine Learning, Mechanistic Interpretability,
  Recurrent Transformers, Hierarchical Reasoning Model, Language Models}

  \vskip 0.3in
]



\printAffiliationsAndNotice{\icmlEqualContribution}

\begin{abstract}
  Mechanistic interpretability seeks to make verifiable statements
  about the internal behavior of large language models (LLMs).
  Many interpretability techniques struggle to
  scale with the increasing size and depth of architectures. Our
  solution to this is to introduce smaller models with structures
  that lend themselves to interpretability.
  In this work, we introduce the Stream Recursion Model (SRM), a
  modification of the Hierarchical Reasoning Model (HRM) designed to
  expose internal computational structure while remaining scalable.
  SRM organizes computation into multiple interacting latent streams
  that are updated through recursive refinement, enabling direct
  analysis of stream dynamics, causal contribution, and routing
  behavior.  SRM achieves performance comparable to
  GPT-2 on a per-parameter basis. Our analysis reveals consistent and
  distinct behavior across streams, indicating structured
  specialization and interaction. These results suggest that SRM
  provides a practical architectural foundation for scalable
  mechanistic interpretability and opens up promising avenues for
  future research in both reasoning performance and interpretability.
\end{abstract}

\section{Introduction}
\label{sec:intro}
The objective of mechanistic interpretability is to analyze the internal
workings of LLMs so as to be able to predict their functionality and
verify their security and stability. The rising use of LLMs in areas
such as healthcare and processing classified documents necessitates
advances in interpretability. Frontier models have recently surpassed
the 1T parameter mark, making mechanistic interpretability difficult
and prohibitively expensive.
The goal of this paper is to develop a more parameter-efficient model
that would structurally lend itself to both new and existing
interpretability techniques.

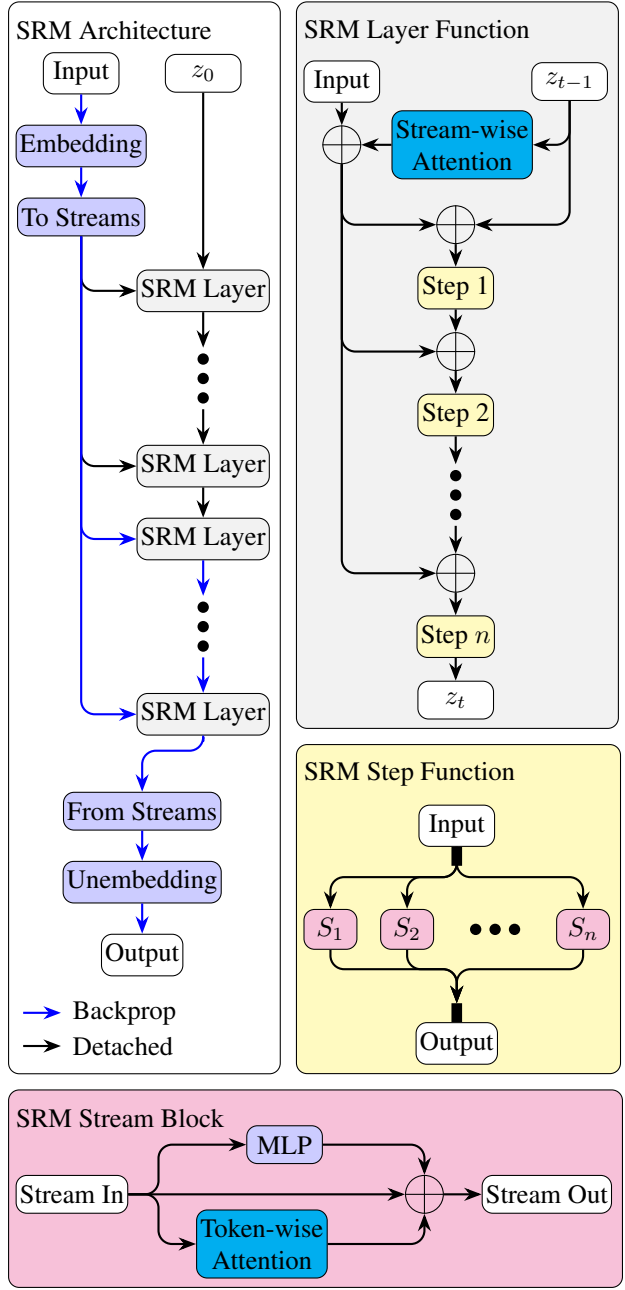
\begin{figure}[!b]
  \centering
  \input{srm-diagrams.tex}
  \caption{Design and layout of the SRM Architecture. stream-wise
    Attention performs attention across the streams per token.
  Token-wise Attention operates like normal attention.}
  \label{fig:SRM}
\end{figure}

\subsection{Related Works}
\label{sec:intro related}

Recent works have shown that there are many redundant parameters
within the layers of vanilla transformer models. In
Mixture-of-Recursions (MoR) \cite{MoR}, looped the middle layers of a
vanilla transformer. They found that in doing so, they were able to
achieve equivalent performance with similar compute time to vanilla
transformers, while using only a third of the
parameters. However, they found that at some point, increasing the
number of looped layers resulted in performance drop-offs. Meanwhile,
\cite{loopedLLM}, found that by looping
through a whole language model multiple times (using a technique
similar to deep supervision from \cite{HRM}) during pre-training,
they were able to outperform models several times larger in complex
reasoning tasks. A drawback with their method is that they were only
able to do 8 loops through the model before they started facing
training issues with unstable gradients, limiting their technique to
only 4 loops.

A promising line of work in parameter reduction emerged from a novel
architecture proposed by \cite{HRM} called the Hierarchical Reasoning
Model (HRM). Combined with a new training technique called deep
reinforcement learning and limited back propagation through time
(LBTT), HRM was able to get $40.3\%$ accuracy on ARC-ARG1 \cite{ARC1}
using only 27M parameters, beating frontier models such as
o3-mini-high with an accuracy of 34.5\%. Building on HRM \cite{TRM}
with their Tiny Reasoning Models (TRM), was able to achieve $44.6\%$
accuracy while using only 7M parameters by increasing parameter reuse
and increasing the depth of the LBTT. The huge performance claims
with a very small number of parameters, along with the highly recurrent
architecture made it very attractive to do interpretability work with.

A novel design element in the HRM architecture is its usage of
multiple latents/streams used in its computation. For consistency in
the rest of this paper we will only refer to them as streams. Both
the HRM and TRM papers propose that 2 streams is the optimal number
of streams. Meanwhile, \cite{SR2} is able to achieve improved
performance using only a single stream with some training tweaks.
\cite{HRM-diffusion} showed that using a single stream with the HRM
architecture resulted in improved performance. This disagreement on
the optimal number of streams suggests that the optimal number of
streams is sensitive to the architectural differences of HRM and its
variants. Furthermore, the inability of these models to improve in
performance when increasing the number of streams, as found by
\cite{TRM}, suggests a flaw in these models.

The field of mechanistic-interpretability has recently been focused
on using dictionary learning techniques. While powerful, dictionary
learning techniques have several drawbacks: they can be more expensive
to train than the original model, are only able to capture features
for a singular point in the model, and have several other open issues
as discussed by \cite{saesurvey}. While Cross Layer Transcoders
(CLTs), another dictionary learning technique, solves some of the
issues SAEs face \cite{CLT}, they are even more expensive to train.

\subsection{Contributions}

In this work we introduce a novel architecture derived from HRM
called the Stream Recursion Model (SRM). We summarize our
contributions as follows:

\begin{itemize}
  \item We introduce a modification to HRM that allows it to scale
    in performance with the use of more streams. This is our SRM model.
  \item We verify that SRM can achieve on-par performance compared to
    vanilla transformers,
    and vastly outperforms HRM when trained on language datasets.
  \item Using techniques made possible by SRM, we identify unique and
    interesting behaviors within our model, such as unique routing
    patterns and specialized behavior of each stream.
\end{itemize}

Furthermore, we achieve the results in this work with only a single
4xL40s GPU cluster over the course of 2 months and $\sim$1000 GPU
hours of external compute.

\section{Motivation}

Based on the disagreements about the best number of streams to use in HRM or
its variants, we believed that there was an issue in how these
models handled communication between streams. Specifically,
streams in HRM communicate information between each other by
simply adding the current state of one stream into another at
specified intervals. This results in the underutilization of each
stream as it does not allow for the translation of information between streams.

While this doesn't appear to be an issue on the non-causal datasets
used by HRM, such as Sudoku \cite{HRM} and ARC-ARG \cite{ARC1,ARC2},
we found that this did cause issues when training on language
datasets. Scaling up HRM's streams in both depth and size was not
able to fix these issues. To solve this issue, we sought to introduce
a learnable translation between streams, as well as increase the
number of streams.

We faced another issue in our attempt to increase the number of
streams. When using 2 streams, the relatively limited number of
possible configurations, and thus parameters, specifying when streams
communicate with each other is feasible to manage. However, the
number of configurations and parameters grows exponentially with the
number of streams. To reasonably use multiple streams, we need a
method which can learn this configuration for us.

We combine both of these solutions - when to communicate and how to
translate between streams - by creating a learnable layer (which we
call the "communication layer") that learns both.

\subsection{The Network Approach}

The Network Approach treats each stream as a node in a network.
Between each node we have a learned translation function
with a learned gating function over the translation. In practical
terms, the network approach is the attention function.

The key difference is that each stream gets its own query, key,
value, and output matrices that it can learn. The value and output
weights work together to create the learned translation between any
two streams. Meanwhile, the key and query weights form our attention
matrix which in turn gives us the gating values for the translation
functions between each pair of streams. By using this approach, we
are still able to view the configuration of communication between the
streams by analyzing the resulting attention matrix.

\section{SRM}
\label{sec:SRM}

In this section we define the Stream Recursion Model. It consists of
3 main functions: the input function, the output function, and the
layer function. The layer function in turn contains 3 functions: a
connection function between the streams, an update function, and a
collection of step-functions
\cref{fig:SRM} along with pseudocode in \cref{fig:srm pseudocode}.
The rest of this section will discuss design decisions and details
for our SRM Model.

\subsection{Input and Output Functions}

A direct embedding from tokens into the streams is inefficient since
the effective embedding width with multiple streams is quite large.
Instead, we embed down to a smaller embedding dimension, then use a
linear layer to get the tokens embedding per stream. We use a similar
method when decoding tokens: a linear of all streams into a small
embedding dimension, followed by an unembedding layer to get
token-ids. By using an intermediate embedding dimension we
drastically reduce the number of parameters used for embedding
and decoding.

\subsection{The Layer Function}

The layer function is the core component that makes SRM work. It
encompasses the connection function, the update function, and the
step functions.

\subsubsection{The Connection Function}

The connection function in SRM is essentially a multi-head- attention
function where each stream learns its own weights for the key, query,
value, and output parameters. Due to this analogy we often refer to
the connection function as stream- wise attention. Pseudocode of the
definition is provided in
\cref{fig:srm pseudocode}.

\subsubsection{The Update Function}

The update function is what connects the step-functions, the
connection function, and the embedded input. It determines how all
three of its variables will interact with each other within the
model, and how gradients will flow during back- propagation. Model
performance is fairly sensitive to the permutation of layernorms used
in the update function. We found that the definition given below
worked the best in our models.

\begin{equation}
  \text{Update}(z_{t,i},c_t,x) = z_{t,i}+\text{RMSNorm}(c_t+x)
  \label{eqn:update}
\end{equation}

Where $z_{t,i}$ is the hidden streams at layer $t$ step $i$, $c_t$ is
the output of the connection function at layer $t$, and $x$ is the
embedding of the input tokens.

\subsubsection{The Step-Function}

The step function consists of a collection transformer blocks, with
each stream having its own unique transformer block. For the
transformer block we utilize a post-norm architecture, following HRM.
In the transformer block we compute the attention function and the
FFN function in parallel, following \cite{gpt-neo} and their GPT-NeoX model.

Each layer in the SRM Model consists of a series of $n$ many steps
and $m$ many step-functions. The layer function hardcodes how its $m$
step functions are called. For example, given the step-functions $A,
B$ and $C$ , one possible hardcoded connection is $ABCABC$; another is $ABCCBA$.

\begin{figure}[!b]
  \centering
  \begin{lstlisting}[language=python]

def streams_linear(z, O):
  B, T, S, E = z.shape # [batch, token, stream, embedding]
  weight = torch.nn.Parameter(S, O, E) # [stream, output, embedding]
  output = torch.einsum('btse,soe->btso', z, weight)
  return output

def streamwise_attention(z, n_heads, head_dim):
  B, T, S, E = z.shape # [batch, token, stream, embedding]

  qkv = streams_linear(z, 3*n_heads*head_dim)
  # [B, T, S*nh, 3*hd]

  qkv = qkv.view(B, T, S, n_heads, 3*head_dim).transpose(3, 2)
  # [B, T, nh, S, 3*hd]

  q,k,v = qkv.split(head_dim, dim=-1)
  # 3 * [B, T, nh, S, hd]

  y = torch.nn.functional.scaled_dot_product_attention(
          q, k, v,
          is_causal=False)
  # [B, T, nh, S, hd]

  y = y.transpose(3,2).view(B, T, n_heads*S, head_dim)
  # [B, T, S*nh, hd]

  y = streams_linear(y, E).view(B, T, S, n_heads, E)
  # [B, T, S*nh, hd] -> [B, T, S*nh, E] -> [B, T, S, nh, E]

  y = y.sum(dim=-2)
  #  [B, T, S, nh, E] -> [B, T, S, E]
  return y

def srm_step(z):
  B, T, S, E = z.shape
  # S x [B, T, E]
  for i in range(S):
    stream = z[:, :, i]
    update = self.mlp[i](stream) + self.attn[i](stream)
    update = RMSNorm(update + stream)
    z[:,:,i] = update
  return z

def srm_layer(x, z, n_steps=3):
  c = streamwise_attention(z)
  y = RMSNorm(c + x)

  for i in range(n_steps):
    step_idx = self.get_step_idx(i)
    z = self.steps[step_idx](z + y)
  return z

def srm(input_ids, z, n, l):
  B, T, S, E = z.shape  # [batch, token, stream, embedding]

  # Input function
  emb = embedding(input_ids) # [B, T, E]
  stream_emb = linear_up(emb).view(B, T, S, E)
  # [B, T, E] -> [B, T, S*E] -> [B, T, S, E]

  for _ in range(n-l):
    with torch.no_grad():
      z = srm_layer(stream_emb, z)
  for _ in range(l):
      z = srm_layer(stream_emb, z)

  # Output function
  out_emb = linear_down(z.view(B, T, S*E))
  # [B, T, S*E] -> [B, T, E]
  output_ids = unembedding(out_emb)

  return output_ids

  \end{lstlisting}
  \caption{Pseudocode for SRM in PyTorch}
  \label{fig:srm pseudocode}
\end{figure}

\section{Hyper-Parameter selection}

This section discusses our efforts in hyperparameter optimization and
selection with HRM models. Due to our limited compute capacity, we
limit our training to a meager 0.5B tokens, or about 1000 training
steps. While this limits the accuracy of the conclusions that we draw
from our experiments, we find that we are still able to determine
which parameters SRM is the most sensitive to, along with a
relatively performant region.

The general hyperparameters that we sought to understand were: (i) the
number of layers used in the model, along with the number of layers
we include during backpropagation, (ii) an interesting modification
involving extra step-functions used during the embedding and
unembedding functions, (iii) the dimension of the streams, along
with their quantity, (iv) and the number of steps and step-functions
within a layer and their ordering.

\subsection{Computation Depth}

Due to the causal nature of language data, we did not implement
Adaptive Computation Time (ACT) \cite{HRM} when training our model on our
language dataset. We did run experiments on the effectiveness of
limited backpropagation through time. Using a fixed depth of 8 layers
we found that returns severely diminished when performing
backpropagation through more than 4 layers.

While we can not be certain that this depth is compute-optimal for
every choice of hyperparameters for SRM, it does provide us a
reasonable choice for us to use throughout the rest of our
experiments and for training. Further work should explore the
sensitivity of performance to the number of layers and
backpropagation layers, with the rest of the hyperparameters for SRM.

\subsection{Pre- and Post-Steps}

Studies on the similarity of layers in dense transformer models has
shown that the first and last layer are significantly different from
the middle layers, suggesting that they are specialized.
\citealp{MoR} demonstrated in their MoR model that adding an
independent transformer at the beginning and end of the model
resulted in significant performance gains. Inspired by these results,
we experimented with adding a step-function after the input function
and before the output function. To our surprise, we found that the
presence of pre-steps resulted in worse performance, while post-steps
did result in either similar or increased performance.

The inclusion of pre-steps may result in lower performance as it
removes the uniformity of token representations leaving the input
function, resulting in less stable gradients throughout the rest of
the model. It is unclear if the presence of post-steps resulted in
performance gains solely due to the increased number of parameters,
or if there is a structural benefit, as would be suggested by the
results found by \cite{MoR}. Because we are uncertain about the
nature of the benefit from post-steps in general, we avoided using them.

\subsection{Layer-Steps}

In order to measure the impact of the number of layer-steps within a
layer and their ordering, we conduct the following experiments: For
each experiment we used three
step-functions $A,B,C$, while we varied both the ordering of the
step-functions and the number of layer-steps. Three was chosen as it
has numerous possible configurations with a minimal amount
of parameters, allowing us more clear insight into the effect of
ordering and layer-step depth. The results are shown in
\cref{fig:step ordering} in Appendix A.

Experimentation revealed that the ordering of the steps didn't have a
noticeable effect and that keeping the number of layer-steps and
step-functions equal gave good performance with the least compute.

\subsection{Stream Size}

To measure the effect of the size of the stream dimension we ran three
sets of experiments. For these experiments we varied 4 variables; the
number of streams $s$, the stream dimension $d_s$, the hidden width
of the step-functions MLP layer $d_h$, and the number of heads used
by the step-functions Attention layer $n_{hs}$.
For the first two experiments we held $s\times d_s = 4096$ constant and
varied $s$. The first set holds parameter count constant by
setting $d_h = 768$ and $n_{hs}=2$. The second set holds the
model width constant by holding $s\times n_{hs}=64$ and $s\times d_h
= 6\times128\times32$. The third set holds stream size constant with
$n_{hs}=2$, $d_h=768$, and $d_s=256$ and measures the impact of more
streams by varying $s$.

As we can see from Figure~\ref{fig:num streams}, there seems to be a
large jump in performance between 4 and 8 streams when the width of
the streams changes. We believe this is due to the dimension of the
streams being greater than the dimension of the intermediate embedding.

\begin{figure}
  \centering
  \begin{tikzpicture}
    \begin{groupplot}[
        group style={
          group size=1 by 2,
          vertical sep=8pt
        },
        ybar,
        width=\linewidth,
        symbolic x coords={1,2,4,8,16,32,64},
        xtick=data,
        tick label style={font=\footnotesize},
        label style={font=\footnotesize},
        legend style={
          font=\footnotesize,
          draw=none,
          at={(0.5,1.12)},
          anchor=south,
          legend columns=3
        }
      ]

      \nextgroupplot[
        bar width=5pt,
        ymin=5.4,
        ymax=7.3,
        height = 4cm,
        ylabel={Loss},
        xticklabels={}
      ]
      \addplot table {data/stream_size/param.dat};
      \addplot table {data/stream_size/width.dat};
      \addplot table {data/stream_size/stream.dat};
      \legend{Fixed Parameter, Fixed Width, Fixed Stream Size}

      \nextgroupplot[
        bar width=5pt,
        ymin=3.6,
        ymax=4.65,
        height = 4cm,
        ylabel={Loss},
        xlabel={Number of Streams}
      ]
      \addplot table {data/stream_size/param.dat};
      \addplot table {data/stream_size/width.dat};
      \addplot table {data/stream_size/stream.dat};

    \end{groupplot}
  \end{tikzpicture}
  ~
  {
    \setlength{\tabcolsep}{4pt}
    \begin{tabular}{|c|c|c|c|c|c|c|c|}
      \cline{2-8}
      \multicolumn{1}{c}{} & \multicolumn{7}{|c|}{Parameter Count (M)}
      \\
      \hline
      \# Streams           & $1$                                       & $2$
      & $4$                                       & $8$   & $16$
      & $32$                                      & $64$                \\
      \hline
      Fixed Parameter      & $85$
      & $85$                                      & $85$
      & $85$                                      & $85$  & $85$ & $85$
      \\
      Fixed Width          & $839$
      & $463$                                     & $261$
      & $161$                                     & $110$ & $85$ & $72$
      \\
      Fixed Stream Size    & 53
      & 54                                        & 56
      & 60                                        & 68    & 85   & 119
      \\
      \hline
    \end{tabular}
  }
  \caption{These hyperparameter sweep experiments were trained for
    1000 steps (approximately 500M tokens) and are not representative
  of the full training runs reported elsewhere.}
  \label{fig:num streams}
\end{figure}

\section{Results}
\label{sec:lang results}

To verify that SRM could perform on language Datasets, we trained 3
models on OpenWebText \cite{OpenWebText}. Model and training details are
provided in \cref{sec:model configuration}.

For a fair comparison, we trained a GPT-2 \cite{gpt2} instance using the same
training configuration as our SRM models. We see that SRM-base is
able to achieve on par performance to GPT-2 with roughly half
the parameters and 7 times the compute time. Meanwhile, SRM-med is
able to out-perform GPT-2 with a similar number of parameters
requiring 6 times more compute. SRM-large demonstrates that SRM is able to
scale well with an increase of parameters.

Of note is the short amount of training steps taken during our
training; we only trained on $\sim$5B tokens. Based on our training
graphs, if we extrapolate out to 300B tokens we expect for SRM-base
to maintain similar performance to GPT-2, SRM-med to stall in
performance, and for SRM-large to reach a loss of around 2.6-2.7.

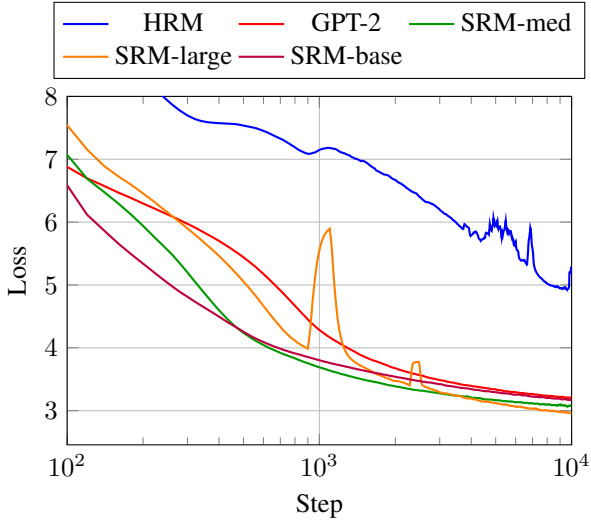
\begin{figure}
  \centering
  \input{training_loss.tex}
  \caption{Training loss of GPT-2 and the three SRM models on
  OpenWebText. Smoothed using a rolling average of 200 steps.}
  \label{fig:training}
\end{figure}

Training results are shown in \cref{table: results}, and a training
graph is shown in \cref{fig:training}.

\begin{table}[h!]
  \centering
  \begin{tabular}{|c|c|c|c| }
    \hline
    Model     & \#Params & Time & Training Loss \\
    \hline
    GPT-2     & 124M     & 11h  & 3.20          \\
    HRM       & 109M     & 22h  & 4.93          \\
    SRM-base  & 68M      & 77h  & 3.18          \\
    SRM-med   & 134M     & 68h  & 3.08          \\
    SRM-large & 454M     & 264h & 2.96          \\
    \hline
  \end{tabular}
  \caption{Performance of GPT-2 and SRM on OpenWebText after 10k
  training steps.}
  \label{table: results}
\end{table}

\input{analysis.tex}

\section{Future Works}

In this section we discuss future directions of study that our SRM
model lends itself to.

\subsection{Mechanistic Interpretability}

The heavily recurrent nature of SRM heavily lends itself to the
universal application of dictionary learning techniques such as
Sparse Autoencoders \cite{saesurvey,SAE} and Cross Layer Transcoders
\cite{CLT}. Combined with our routing analysis, one could use
dictionary learning to discover if routing switches are based on some
interpretable feature. We leave this direction of study to future work.

\subsection{Reasoning Tasks}

Because SRM is a variant of HRM, we believe that it is well suited
to logic and reasoning tasks. Further work training SRM on language
based reasoning tasks could yield substantial results if deep
reinforcement learning \cite{HRM} is used.

\section{Conclusion}
In this work we demonstrated that our proposed SRM architecture both
generalizes and improves upon the HRM architecture. By allowing
for a learned communication method between streams with our
connection function, we can generalize HRM to use multiple streams.
This allows for expanded capability, along with opening up new avenues
of analysis for mechanistic interpretability.

We demonstrated that HRM, as it was originally designed
was incapable of performing on language datasets. We show
that SRM is capable of performing comparably to standard
transformer models such as GPT-2.

In our analysis of our SRM models we found that different
choices in sizes and parameters yields significantly distinct and
interesting routing structures. This result provides
insights on the role and purposes of the different streams. Future
work performing in-depth studies on the individual streams may
identify critical roles and functionality in the individual streams
which may mimic the region of the human brain.

\section*{Acknowledgements}

Sandia National Laboratories is a multimission laboratory managed and
operated by National Technology \& Engineering Solutions of Sandia,
LLC, a wholly owned subsidiary of Honeywell International Inc., for
the U.S. Department of Energy’s National Nuclear Security
Administration under contract DE-NA0003525. SAND2026-21799O.

This paper describes objective technical results and analysis. Any
subjective views or opinions that might be expressed in the paper do
not necessarily represent the views of the U.S. Department of Energy
or the United States Government.

\bibliography{sources}
\bibliographystyle{icml2026}

\newpage
\appendix
\onecolumn
\section{Additional Tables and Results}
\begin{figure}[h!]
  \centering
  \begin{tabular}{|l|c|}
    \hline
    Step Ordering          & Loss   \\
    \hline
    $ABC$                  & $3.84$ \\
    \hline
    $ABCA$                 & $3.84$ \\
    \hline
    $ABCABC$               & $3.83$ \\
    \hline
    $ABCABCABC$            & $3.82$ \\
    \hline
    $ABBC$                 & $3.83$ \\
    \hline
    $ABBBBC$               & $3.82$ \\
    \hline
    Random($\{A,B,C\}, 6$) & $3.87$ \\
    \hline
  \end{tabular}
  \caption{Effect on the ordering of the step-functions on loss after
    1000 training steps. Random indicates that for each step we
    randomly choose one of the three step-functions, and that each
  layer had 6 layer steps.}
  \label{fig:step ordering}
\end{figure}

\section{Model configuration and training}
\label{sec:model configuration}

All trained models utilize the Adam-atan2 \cite{adamatan2} optimizer
they also use the tokenizer from GPT-2 \cite{gpt2}.
The SRM models use the Network approach for their connection function. We use
a head dimension of 64 for all attention modules. The embedding
dimension and the stream dimension are both 512. For the streams we
use GELU activation with a
hidden dimension of 3072 for the MLP layer, along with 8 heads for the tokenwise
attention. We use a model depth of 8 layers, performing
backpropagation through the last 4.

For training, our batch size during training is 512, and we train for a
total of 10k steps, resulting in using $\sim5$B training tokens. We
use a learning rate of $1e-3$, and a weight decay of $0.1$.

SRM-med has 8 streams, uses 4 heads for the
stream-wise-attention, and has 3 layer steps with 3 step-functions.
Following \cite{HRM}, it uses 200 warm-up steps during training.

SRM-large has 16 streams, uses 8 heads for the
stream-wise-attention, and has 6 layer steps with 6 step-functions. We
used 1000 warm-up steps during training.

SRM-base was trained using 32 streams, 6 layer steps with 3
step-functions, 4 stream-wise-attention heads, a stream dimension of
128 with 2 heads for token-wise attention and a hidden dimension of
768 for the MLP layer. A single post-step is used. We only used 100
warm-up steps during training.

GPT-2 was trained without any warm-up steps using the default model
configuration for GPT2LMHead provided by Huggingface's transformers
python package.

HRM was trained with an embedding dimension of 768. For the MLP
layers, we used with GELU activation with a hidden dimension of 3072.
For MHA, we used a head dimension of 64 and 12 heads. For HRM
specific parameters, the L-net had a depth of 6 transformer blocks,
while the H-net was 4 blocks deep. We used 3 H-cycles and 4 L-cycles.
During training, we used a step size of $1e-4$, and used 1000 warm-up steps.

\input{analysis_extra.tex}

\end{document}

%% file: srm-diagrams.tex
\tikzset{
  Label/.style={
    align=center
  },
  EndBlock/.style={
    draw,
    rounded corners,
    minimum width=10mm,
    minimum height=5mm,
    align=center,
    fill=white
  },
  Blank/.style={
    minimum width=0mm,
    minimum height=0mm,
    inner sep=0mm
  },
  LayerFunctionBlock/.style={
    draw,
    rounded corners,
    minimum width=10mm,
    minimum height=5mm,
    align=center,
    fill=gray!10
  },
  StepFunction/.style={
    draw,
    rounded corners,
    minimum width=10mm,
    minimum height=5mm,
    align=center,
    fill=yellow!30
  },
  ConnectionFunction/.style={
    draw,
    rounded corners,
    minimum width=10mm,
    minimum height=5mm,
    align=center,
    fill=red!20
  },
  InnerConnectionFunction/.style={
    draw,
    rounded corners,
    minimum width=10mm,
    minimum height=5mm,
    align=center,
    fill=red!10
  },
  StreamBlock/.style={
    draw,
    rounded corners,
    minimum width=7mm,
    minimum height=5mm,
    align=center,
    fill=magenta!30
  },
  MultiLayerPerceptron/.style={
    draw,
    rounded corners,
    minimum width=10mm,
    minimum height=5mm,
    align=center,
    fill=blue!20
  },
  AttentionBlock/.style={
    draw,
    rounded corners,
    minimum width=10mm,
    minimum height=5mm,
    align=center,
    fill=cyan!100
  },
  SumBlock/.style={
    draw,
    circle,
    append after command={
      \pgfextra{\let\LN\tikzlastnode}
      (\LN.north) edge (\LN.south)
      (\LN.west) edge (\LN.east)
    },
    minimum size=5mm,
    inner sep=0pt,
    font=\Large
  },
  VDot/.style={
    append after command={
      \pgfextra{\let\LN\tikzlastnode}
      node[circle, fill=black, minimum size = 1.5mm, inner sep=0pt,
      yshift=-1.5mm] at (\LN.north) {}
      node[circle, fill=black, minimum size = 1.5mm, inner sep=0pt,
      yshift=0mm] at (\LN.center) {}
      node[circle, fill=black, minimum size = 1.5mm, inner sep=0pt,
      yshift=1.5mm] at (\LN.south) {}
    },
    minimum size=8mm,
    inner sep=0pt,
    align=center
  },
  HDot/.style={
    append after command={
      \pgfextra{\let\LN\tikzlastnode}
      node[circle, fill=black, minimum size = 1.5mm, inner sep=0pt,
      xshift=-1.5mm] at (\LN.east) {}
      node[circle, fill=black, minimum size = 1.5mm, inner sep=0pt,
      xshift=0mm] at (\LN.center) {}
      node[circle, fill=black, minimum size = 1.5mm, inner sep=0pt,
      xshift=1.5mm] at (\LN.west) {}
    },
    minimum size=8mm,
    inner sep=0pt,
    align=center
  },
  flow/.style={
    ->,
    thick,
    >=Stealth,
    rounded corners=1.8mm
  },
  bflow/.style={
    ->,
    thick,
    draw=blue,
    >=Stealth,
    rounded corners=1.8mm
  },
  dflow/.style={
    thick,
    dotted,
    rounded corners=1.8mm
  },
  tflow/.style={
    line width=4pt,
    rounded corners=1.8mm
  }
}

\begin{tikzpicture}[node distance=2mm]
  \node[EndBlock, inner xsep=0.5mm, anchor=north west] (srmArchitecture) {
    \begin{tikzpicture}[node distance=4mm, every node/.style={anchor=center}]
      \node[Label] (label) {SRM Architecture};

      \node[MultiLayerPerceptron, anchor=north west] (embedding) at
      ($(label.south west) + (0.5mm,-10mm)$) {Embedding};
      \node[EndBlock, above=of embedding] (input) {Input};
      \node[EndBlock, right=of input, xshift=2mm] (zNot) {$z_0$};

      \node[MultiLayerPerceptron, below=of embedding] (toStreams) {To Streams};
      \node[LayerFunctionBlock] (srmLayer1) at ($(zNot.center |-
      toStreams.south) + (0mm,-7.2mm)$) {SRM Layer};
      \node[VDot, below=of srmLayer1, yshift=-2] (vdot1) {};
      \node[LayerFunctionBlock, below=of vdot1, yshift=-2]
      (srmLayer2) {SRM Layer};
      \node[LayerFunctionBlock, below=of srmLayer2] (srmLayer3) {SRM Layer};
      \node[VDot, below=of srmLayer3, yshift=-2] (vdot2) {};
      \node[LayerFunctionBlock, below=of vdot2, yshift=-2]
      (srmLayer4) {SRM Layer};
      \coordinate (p1) at ($(srmLayer4.south)-(0mm,2mm)$);

      \coordinate (mid) at ($(input.center)!0.5!(zNot.center)$);
      \coordinate (target) at ($(srmLayer4.south)+(0mm,-10mm)$);
      \node[MultiLayerPerceptron] (fromStreams) at ($(mid |-
      target)$) {From Streams};
      \node[MultiLayerPerceptron, below=of fromStreams] (unembedding)
      {Unembedding};
      \node[EndBlock, below=of unembedding] (output) {Output};

      \draw[bflow] (input) -- (embedding);
      \draw[bflow] (embedding) -- (toStreams);
      \draw[flow] (zNot) -- (srmLayer1);
      \draw[flow] (toStreams) |- (srmLayer1);
      \draw[flow] (toStreams) |- (srmLayer2);
      \draw[bflow] (toStreams) |- (srmLayer3);
      \draw[bflow] (toStreams) |- (srmLayer4);
      \draw[flow] (srmLayer1) -- (vdot1);
      \draw[flow] (vdot1) -- (srmLayer2);
      \draw[flow] (srmLayer2) -- (srmLayer3);
      \draw[bflow] (srmLayer3) -- (vdot2);
      \draw[bflow] (vdot2) -- (srmLayer4);
      \draw[bflow] (srmLayer4) |- (p1) -| (fromStreams);
      \draw[bflow] (fromStreams) -- (unembedding);
      \draw[bflow] (unembedding) -- (output);

      \node[Label, anchor=north west] (backprop) at ($(embedding.west
      |- output.south) + (7mm,-2mm)$) {Backprop};
      \coordinate (backpropStart) at ($(backprop.west) + (-6mm,0mm)$);
      \coordinate (backpropEnd) at ($(backprop.west) + (-1mm,0mm)$);
      \draw[bflow] (backpropStart) -- (backpropEnd);

      \node[Label, anchor=north west] (detached) at ($(backprop.south
      west) + (0mm,1mm)$) {Detached};
      \coordinate (detachedStart) at ($(detached.west) + (-6mm,0mm)$);
      \coordinate (detachedEnd) at ($(detached.west) + (-1mm,0mm)$);
      \draw[flow] (detachedStart) -- (detachedEnd);
    \end{tikzpicture}
  };
  \node[LayerFunctionBlock, inner xsep=0.5mm, anchor=north west]
  (srmLayerFunction) at ($(srmArchitecture.north east) + (2mm,0mm)$) {
    \begin{tikzpicture}[node distance=3mm, every node/.style={anchor=center}]
      \node[Label] (label) {SRM Layer Function};
      \node[EndBlock, anchor=north west] (input) at ($(label.south
      west) + (0.5mm,-1mm)$) {Input};
      \node[EndBlock, right=of input, xshift=17mm] (zIsub1) {$z_{t-1}$};
      \node[SumBlock, below=of input] (sum0) {};
      \node[AttentionBlock, right=of sum0, xshift=1mm]
      (connection) {Stream-wise\\Attention};

      \coordinate (mid) at ($(input.center)!0.5!(zIsub1.center)$);
      \coordinate (target) at ($(connection.south)+(0mm,-6mm)$);
      \node[SumBlock] (sum1) at (mid |- target) {};
      \node[StepFunction, below=of sum1] (step1) {Step 1};
      \node[SumBlock, below=of step1] (sum2) {};
      \node[StepFunction, below=of sum2] (step2) {Step 2};
      \node[VDot, below=of step2, yshift=-2] (vdot1) {};
      \node[SumBlock, below=of vdot1, yshift=-2] (sumn) {};
      \node[StepFunction, below=of sumn] (stepn) {Step $n$};
      \node[EndBlock, below=of stepn] (zt) {$z_t$};

      \draw[flow] (input) -- (sum0);
      \draw[flow] (zIsub1) |- (connection);
      \draw[flow] (connection) -- (sum0);
      \draw[flow] (zIsub1) |- (sum1);
      \draw[flow] (sum0) |- (sum1);

      \draw[flow] (sum1) -- (step1);
      \draw[flow] (step1) -- (sum2);
      \draw[flow] (sum0) |- (sum2);
      \draw[flow] (sum2) -- (step2);
      \draw[flow] (step2) -- (vdot1);
      \draw[flow] (vdot1) -- (sumn);
      \draw[flow] (sum0) |- (sumn);
      \draw[flow] (sumn) -- (stepn);
      \draw[flow] (stepn) -- (zt);
    \end{tikzpicture}
  };
  \node[StepFunction, inner xsep=0.5mm, anchor=north west]
  (srmStepFunction) at ($(srmLayerFunction.south west) + (0mm,-2mm)$) {
    \begin{tikzpicture}[node distance=3mm, every node/.style={anchor=center}]
      \node[Label] (label) {SRM Step Function};
      \node[StreamBlock, anchor=north west] (S1) at ($(label.south
      west) + (0.5mm,-15mm)$) {$S_1$};
      \node[StreamBlock, right=of S1] (S2) {$S_2$};
      \node[HDot, right=of S2, xshift=1mm] (hdot) {};
      \node[StreamBlock, right=of hdot, xshift=1mm] (Sn) {$S_n$};

      \node[EndBlock] (input) at ($(S1.north)!0.5!(Sn.north) +
      (0mm,11mm)$) {Input};
      \coordinate (thickStart) at ($(input.south) + (0mm,-2.5mm)$);
      \coordinate (split) at ($(input.south) + (0mm,-4mm)$);
      \node[EndBlock] (output) at ($(input.south) + (0mm,-26mm)$) {Output};
      \coordinate (thickEnd) at ($(output.north) + (0mm,2.5mm)$);
      \coordinate (join) at ($(output.north) + (0mm,7mm)$);

      \draw[tflow] (input) -- (thickStart);
      \draw[flow] (input) -- (split) -| (S1);
      \draw[flow] (input) -- (split) -| (S2);
      \draw[flow] (input) -- (split) -| (Sn);

      \draw[tflow] (thickEnd) -- (output);
      \draw[flow] (S1) |- (join) -- (thickEnd);
      \draw[flow] (S2) |- (join) -- (thickEnd);
      \draw[flow] (Sn) |- (join) -- (thickEnd);

    \end{tikzpicture}
  };

  \node[StreamBlock, inner xsep=0.5mm, anchor=north west]
  (srmStreamBlock) at ($(srmArchitecture.south west) + (0mm,-2mm)$) {
    \begin{tikzpicture}[node distance=3mm, every node/.style={anchor=center}]
      \node[Label] (label) {SRM Stream Block};
      \node[EndBlock, anchor=north west] (input) at ($(label.south
      west) + (0.5mm,-5mm)$) {Stream In};
      \node[EndBlock, anchor=east, right=of input, xshift=43.75mm]
      (output) {Stream Out};

      \node[MultiLayerPerceptron] (mlp) at
      ($(input.north)!0.4!(output.north) + (3mm,4mm)$) {MLP};
      \node[AttentionBlock] (attention) at
      ($(input.south)!0.4!(output.south) + (0mm,-4mm)$) {Token-wise\\Attention};
      \node[SumBlock, left=of output, xshift=-2mm] (sum) {};

      \draw[flow] (input) -- (sum);
      \coordinate (p1) at ($(input.east) + (4mm,0mm)$);
      \draw[flow] (input) -- (p1) |- (mlp);
      \draw[flow] (input) -- (p1) |- (attention);

      \draw[flow] (mlp) -| (sum);
      \draw[flow] (attention) -| (sum);
      \draw[flow] (sum) -- (output);
    \end{tikzpicture}
  };

\end{tikzpicture}

%% file: training_loss.tex
\begin{tikzpicture}
  \begin{axis}[
      width=\columnwidth,
      height=0.75\columnwidth,
      xlabel={Step},
      ylabel={Loss},
      xmode=log,
      xmin=100,
      xmax=10000,
      ymax=8,
      ytick distance=1,
      grid=major,
      legend style={
        at={(0.5,1.05)},
        anchor=south,
        legend columns=3,
      },
    ]

    \addplot[blue, thick] table[
      x=Step,
      y={HRM - loss},
      col sep=comma,
    ] {data/loss_smoothed.csv};
    \addlegendentry{HRM}

    \addplot[red, thick] table[
      x=Step,
      y={gpt2-test - loss},
      col sep=comma,
    ] {data/loss_smoothed.csv};
    \addlegendentry{GPT-2}

    \addplot[green!60!black, thick] table[
      x=Step,
      y={161M parameters - loss},
      col sep=comma,
    ] {data/loss_smoothed.csv};
    \addlegendentry{SRM-med}

    \addplot[orange, thick] table[
      x=Step,
      y={479M parameters - loss},
      col sep=comma,
    ] {data/loss_smoothed.csv};
    \addlegendentry{SRM-large}

    \addplot[purple, thick] table[
      x=Step,
      y={4-Bl_3-Pg_6-Ls_1-Ps - loss},
      col sep=comma,
    ] {data/loss_smoothed.csv};
    \addlegendentry{SRM-base}

  \end{axis}
\end{tikzpicture}

%% file: analysis.tex
\section{Analysis}

In this section we study stream level behavior by performing the 3
following experiments.

\begin{enumerate}
  \item Routing Behaviors
  \item Connection Update Behavior Alignment
  \item Step-Function Update Behavior Alignment
\end{enumerate}

We perform these studies across three different SRM Configurations of
varying sizes, SRM-base, SRM-med, and SRM-large. Model configurations
are provided in \cref{sec:model configuration}. Further details and
analysis, along with additional experiments, are performed and outlined in
\cref{sec:appendix_baseline,sec:appendix_161m,sec:appendix_479m}.

\subsection{Routing Behaviors}

The SRM's stream-wise attention mechanism
enables each stream to selectively communicate with other streams,
creating a learned routing network. We analyze these routing patterns
to characterize the model's internal communication structure. We
analyze edges whose value exceeds $0.10$ in the attention matrix
after applying softmax; this gives us edges of strong significance.

In SRM-base, we observe a small subset of streams consistently
exhibiting higher outgoing routing frequency. These streams route to
a broader set of streams than the rest, indicating a hub-like
topology. The routing patterns are
stable across layers and attention heads, suggesting that routing
patterns occur uniformly across model depth. See \cref{fig:baseline_routing}.

In SRM-med, like in SRM-base, we observe a
small subset of streams that tends to frequently output into the rest of
the other streams. Of particular note with SRM-med is the low self
routing frequency for the non-hub-like streams. See \cref{fig:161m_routing}.

\begin{figure*}[!htb]
  \centering
  \includegraphics[width=1.9\columnwidth]{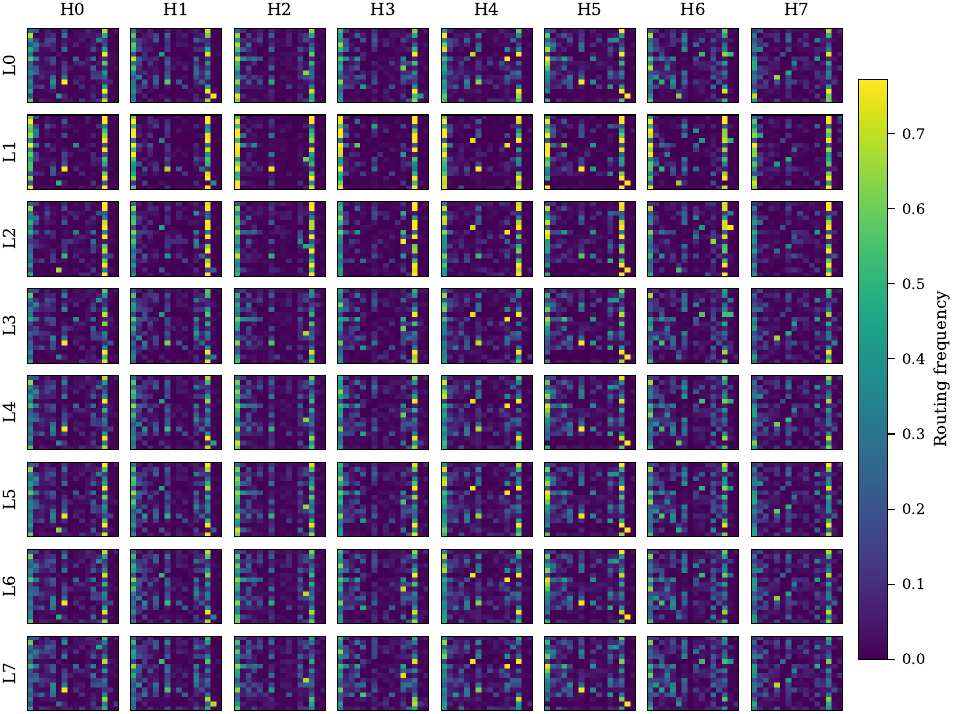}
  \caption{Stream-to-stream routing frequency for SRM-large at a 10\%
    attention threshold. The x-axis across heat-maps is the source
  stream, while the y-axis is the target stream.   }
  \label{fig:479m_routing2}
\end{figure*}

In SRM-large, we once again observe a small subset of hub-like
streams. Unlike SRM-base, we are able to easily observe that there is
a subset of streams that act as sinks, rarely outputting into
other streams. We also see that there is a noticeable change in
routing behavior throughout the models depth. See \cref{fig:479m_routing2}.

It is interesting to observe that by adding more streams, significant
routing becomes more sparse under the threshold. Hub-like streams
continue to be observed, and the number of seemingly specialized
connections and streams increases.

\subsection{Connection Update Behavior Alignment}
Here we measure the
cosine similarity of the output of the connection function $c_t$ and
the token input $x_t$ in the update function.

Across all three models we observe that $c_t$ and $x_t$ are nearly
orthogonal, with a small positive bias in cosine similarity. This
suggests that the connection function introduces directionally
distinct information to the streams, rather than reinforcing input
token direction or canceling it out. See
\cref{fig:baseline_update_attn,fig:161m_update_attn,fig:479m_update_attn}.

\subsection{Step-Function Update Behavior}
When comparing $\text{RMSNorm}(c_t+x)$ and the output of the
step-function, all three models have slightly different behaviors. In
SRM-base, all streams maintain a relatively high cosine similarity
$>0.7$, with the exception of a few streams that hover around a
cosine similarity
between $0.0$ and $0.3$, see \cref{fig:baseline_update_state}. In
SRM-med, we observe that all streams maintain a high cosine
similarity $>0.5$, see \cref{fig:161m_update_state}.

SRM-large exhibits unique behavior compared to the other models.
While most streams maintain a cosine similarity $>0.5$, one stream
hovers around $0.0$, except for the last step in the layer, where it
jumps to $>0.8$. Two other streams display distinct behavior,
maintaining negative values ranging from $-0.1$ to $-0.7$, except for
the last step in the layer, where similarity hovers around $0.3$, see
\cref{fig:479m_update_state}. These behaviors suggest that
streams in SRM-large are performing a functionally different roles
compared to the other streams and the other models.

%% file: analysis_extra.tex
\section{SRM-base: Experiments and Results}
\label{sec:appendix_baseline}

This appendix provides detailed experimental results for the baseline
SRM model (32 streams, 128 dimensions each, ABCABC step-group ordering).

\subsection{Representational Drift (KL Divergence)}

For the baseline SRM with 128-dimensional streams, per-stream KL
divergence between consecutive layer-wise step-group contributions
provides a stable and interpretable measure of representational
change. This metric quantifies how much a stream's representation
changes from one layer to the next for a given step group. Unlike the
SRM-med and SRM-large models where KL divergence becomes numerically unstable
due to higher-dimensional streams (512 dimensions), the baseline
model's lower-dimensional streams yield reliable KL measurements.

Across streams and layers, KL values show substantial variation. Some
streams exhibit consistently low divergence, indicating stable or
slowly evolving representations, while others display sharp spikes at
specific depths, suggesting transformation events. Convergence
analysis, defined as the earliest transition after which KL changes
by less than 10\% for the remainder of the forward pass, reveals
three categories of behavior: early, middle, and late convergers.

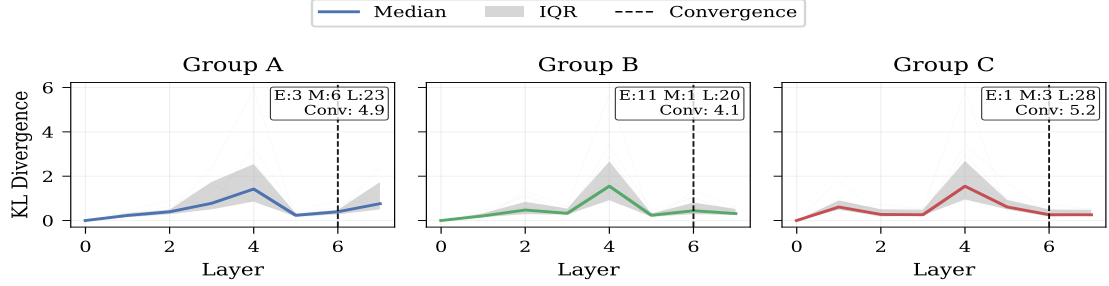
\begin{figure}[H]
  \centering
  \input{figs/baseline_kl_three_panel.tex}
  \caption{Per-stream KL divergence measured between consecutive
    layer-wise step-group contributions for the baseline SRM (ordering:
    ABCABC). Solid lines show the median KL divergence across streams;
    shaded regions indicate the interquartile range (25--75\%). Faint
    trajectories correspond to individual streams. The dashed vertical
    line marks the median convergence transition. KL divergence provides
    a stable and interpretable measure of representational change for the
  baseline model.}
  \label{fig:baseline_kl}
\end{figure}

A pronounced Layer-4 transition spike appears across all step groups,
indicating a global structural transition in the SRM's recursive
computation rather than a group-specific effect. Overall, the
baseline results demonstrate clear functional differentiation among
streams, motivating deeper analysis at larger scales.

\subsection{Stream Ablation}

Mean ablation reveals a strongly right-skewed distribution of stream
importance. Most streams are associated with small KL changes when
ablated, while a small subset produces substantially larger
degradation in next-token prediction.

The most impactful stream produces a KL divergence of approximately
0.091, while the least impactful stream produces approximately 0.009,
representing an order-of-magnitude difference in causal contribution.
A small group of streams (notably streams 16, 26, 17, and 21) form a
clear high-importance cluster, standing well above the remainder of
the distribution.

A cumulative contribution analysis further highlights this hierarchy.
The top nine streams account for approximately 50\% of the total
causal impact, the top sixteen account for roughly 75\%, and the top
nineteen account for approximately 80\%. The remaining streams
contribute relatively little individually, indicating diminishing
marginal impact beyond the most influential subset.

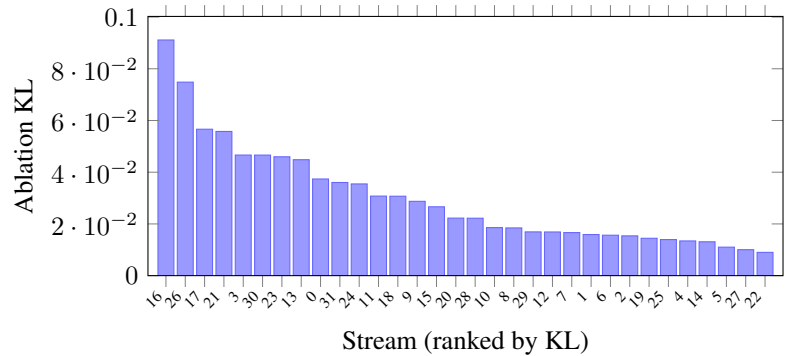
\begin{figure}[H]
  \centering
  \input{figs/baseline_ablation.tex}
  \caption{Per-stream ablation impact for the SRM-base model, ranked by
    KL divergence. The top nine streams account for approximately half of
    the total causal impact, while all streams contribute non-negligibly
  to next-token prediction.}
  \label{fig:baseline_ablation}
\end{figure}

\subsection{Update Alignment}

\subsubsection{Attention-Token Alignment}

In the baseline SRM, cosine similarity between attention output $c_t$
and token input $x_t$ is consistently near zero across streams and
layers (mean $\approx$ 0.067, range -0.070 to 0.175). Approximately
10.5\% of stream-layer pairs exhibit negative cosine values,
indicating that attention output can point in the opposite direction
of token input.

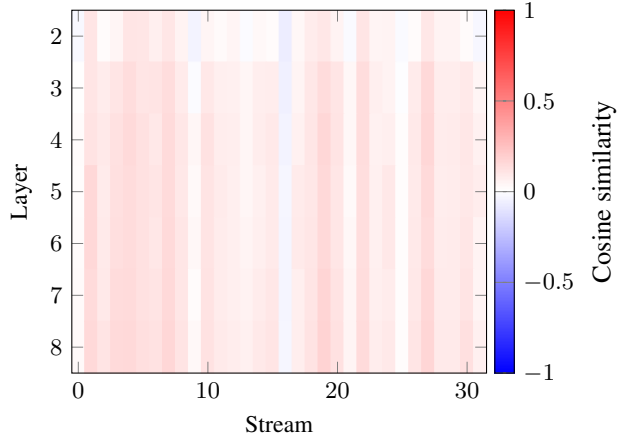
\begin{figure}[!htb]
  \centering
  \input{figs/baseline_update_attn.tex}
  \caption{Cosine similarity between attention output and token input
    for the SRM-base model. Values cluster near zero across all layers and
    streams, indicating that attention-derived updates are largely
  orthogonal to token inputs.}
  \label{fig:baseline_update_attn}
\end{figure}

High-impact streams identified in the ablation analysis exhibit
similar behavior. For example, Stream 16 (highest ablation KL) shows
negative mean alignment (-0.040), while Streams 26, 17, and 21 show
small positive values (0.024-0.068). These results indicate that
streamwise attention does not simply amplify token representations,
but instead introduces directionally distinct information.

\subsubsection{Update-State Alignment}

In the baseline SRM, cosine similarity between the RMS-normalized
combined update $(c_t + x_t)$ and the RMS-normalized hidden stream
state is strongly positive across all streams and layers. No
systematic negative alignment is observed. Alignment values remain
high and stable across recursive steps within each layer, indicating
that updates are directionally consistent with the existing stream state.

Across layers, alignment exhibits clear stepwise structure
corresponding to the six recursive steps per layer. Within each
layer, alignment remains consistently positive across steps. Minor
variation across streams is observed but remains bounded and does not
disrupt the overall alignment pattern.

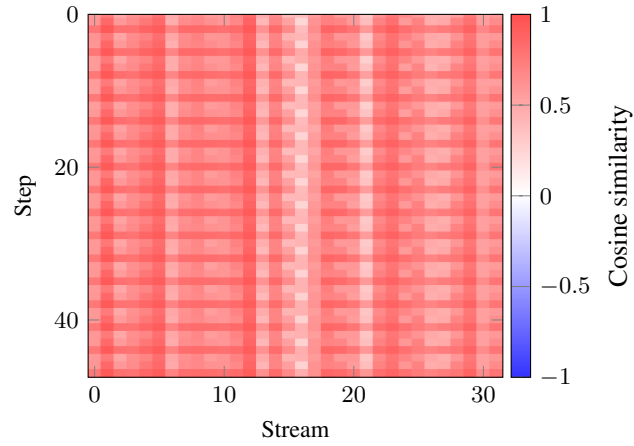
\begin{figure}[!htb]
  \centering
  \input{figs/baseline_update_state.tex}
  \caption{Cosine similarity between the combined update and the
    existing stream state for the SRM-base model. Strongly positive
    alignment across layers and streams indicates that updates reinforce
  rather than oppose current representations.}
  \label{fig:baseline_update_state}
\end{figure}

\subsection{Routing Structure}

In the baseline SRM, routing activity is unevenly distributed across
streams. A subset of streams consistently exhibits higher outgoing
routing frequency, routing to a broader set of target streams than
others. This pattern is stable across layers (L0-L7) and attention
heads, indicating that outgoing routing activity does not vary
substantially with depth.

Routing events exceeding the 10\% threshold are not uniformly
distributed. Instead, they are concentrated in specific source
streams, while many streams rarely exceed the threshold as routing
origins. This concentration persists across layers, suggesting that
the stream-to-stream routing structure is established early and
maintained through depth.

As a reference point, under a uniform attention distribution with 32
streams, the expected attention weight assigned to any individual
stream would be approximately 3.1\%. The observed routing frequencies
therefore reflect attention patterns that substantially exceed
uniform allocation.

\begin{figure}[!htb]
  \centering
  \input{figs/baseline_routing.tex}
  \caption{Stream-to-stream routing frequency for the SRM-base model at
    a 10\% attention threshold. Rows correspond to layers and columns to
    attention heads. Each cell indicates routing from source (y) to
    target (x). Patterns are dense and broadly distributed across the
  network.}
  \label{fig:baseline_routing}
\end{figure}
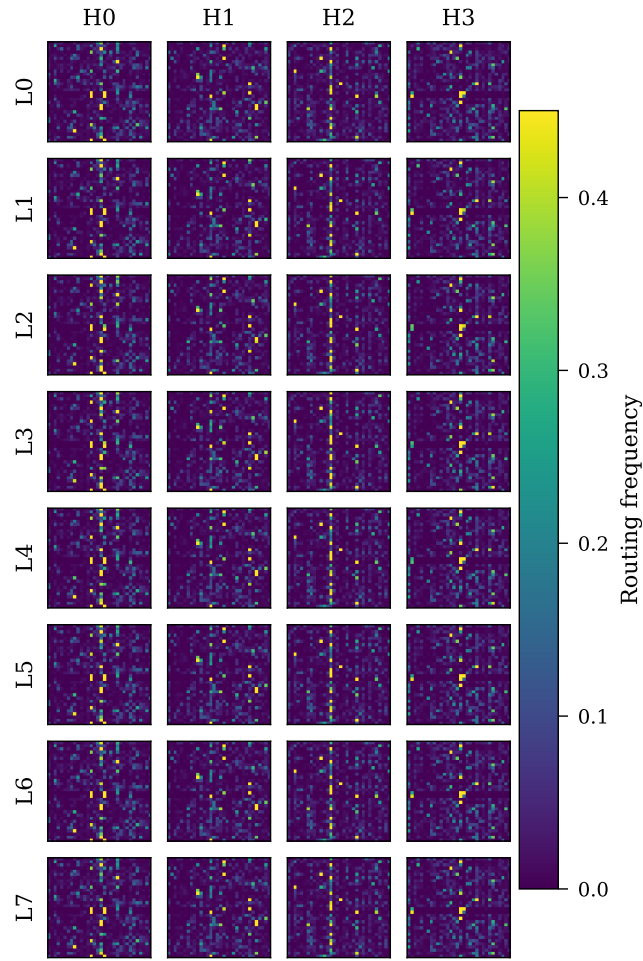

\subsection{Stream-wise Logit Lens}

In the baseline SRM, no individual stream provides a reliable
standalone predictor of the next token. Absolute predictive
performance remains weak across all streams and layers, confirming
that prediction is not localized to any single stream representation.

Despite this, the logit lens reveals a strong depth-dependent
structure shared across streams. All streams follow a similar
trajectory: predictive alignment improves early, degrades sharply at
mid-layers, and then partially recovers toward later layers. This
pattern is consistent across streams, indicating that the effect is
driven by layer-wise computation rather than stream-specific behavior.

The coordinated drop in predictive quality at mid-depth indicates a
phase where stream representations temporarily move away from the
output-aligned subspace. This is followed by a gradual recovery in
later layers. Importantly, streams do not separate into distinct
predictive trajectories during this phase.

\begin{figure}[!htb]
  \centering
  \input{figs/baseline_logit_curves.tex}
  \caption{Per-stream logit lens curves (mean-corrected) for
    SRM-base. Each panel shows the mean log-probability of the correct
  token across layers for an individual stream.}
  \label{fig:baseline_logit_curves}
\end{figure}
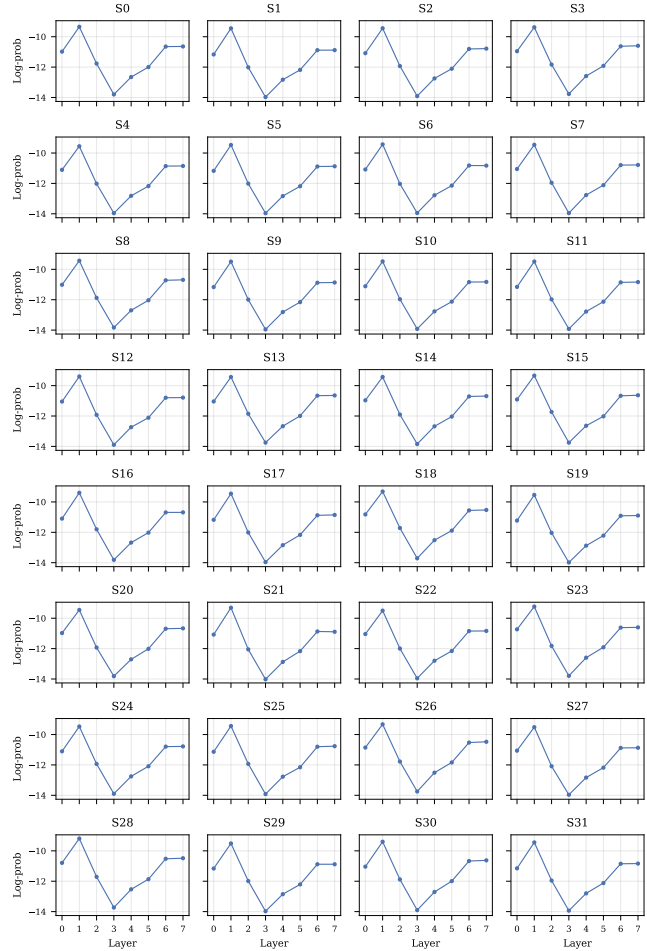

\begin{figure}[!htb]
  \centering
  \input{figs/baseline_logit_heatmap.tex}
  \caption{Stream-wise logit lens (mean-corrected) for the SRM-base model.
    The heatmap shows the mean log-probability of the correct token across
    layers for each stream. A coordinated mid-layer degradation is visible
  across nearly all streams, followed by partial recovery at later layers.}
  \label{fig:baseline_logit_heatmap}
\end{figure}
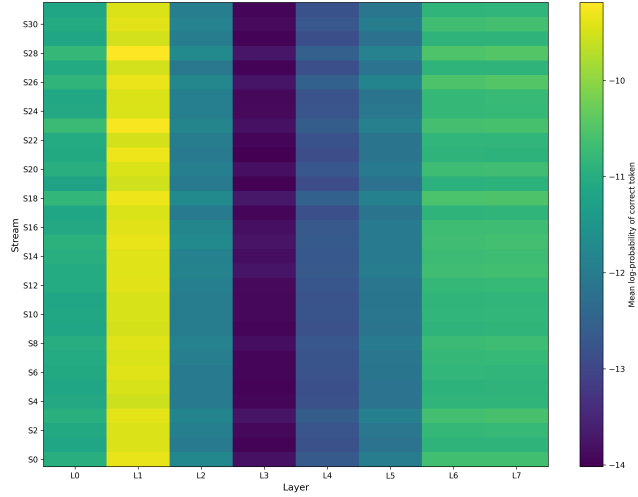

The heatmap representation (Figure~\ref{fig:baseline_logit_heatmap})
confirms this pattern, showing the coordinated mid-layer degradation
visible across nearly all streams, followed by partial recovery at later layers.

\section{SRM-med Experiments and Results}
\label{sec:appendix_161m}

This appendix provides detailed experimental results for the SRM-med SRM
model (8 streams, 512 dimensions each, ABC step-group ordering).

\subsection{Representational Drift (Cosine Distance)}

In the SRM-med model, streams have 512-dimensional representations. When
applying the baseline KL formulation to these high-dimensional
updates, KL divergence becomes numerically unstable, producing
extreme spikes and NaN/Infinity values. We therefore use cosine
distance drift as the primary metric for this model.

The three step groups (A, B, C) exhibit systematically different
drift and convergence profiles based on cosine distance. Groups A and
B show persistent non-zero drift across most layer transitions, with
all streams in both groups converging late at mean convergence of
6.0 $\pm$ 0.0 layers (median: transition 6, L6$\rightarrow$L7). This
indicates that representations associated with these step groups
continue to change until the final layer transitions. In contrast,
Group C shows earlier stabilization for a subset of streams:
specifically, 5 streams converge by transition 2 (L2$\rightarrow$L3),
while 3 streams remain late convergers, yielding mean convergence of
3.5 $\pm$ 2.1 layers (median: transition 2). Group C also exhibits
larger drift magnitudes at early transitions, followed by reduced
drift at later depths.

\begin{figure}[!htb]
  \centering
  \input{figs/161m_drift.tex}
  \caption{Per-stream representational drift measured by cosine distance across depth for the SRM-med model. Colored lines denote the median across streams, with shaded regions indicating the interquartile range. A pronounced mid-depth transition is visible across step groups, indicating coordinated representational change.}
  \label{fig:161m_drift}
\end{figure}
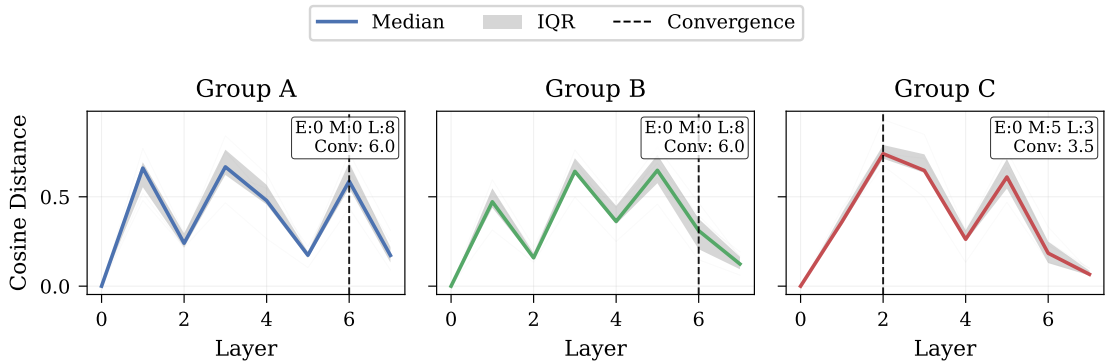

\subsection{Stream Ablation}

The SRM-med model exhibits a similar but more compact hierarchy due to
the smaller number of streams. Ablation KL values span a dynamic
range of 0.130 to 0.339 bits ($\approx$2.6$\times$), with a mean of
0.224 $\pm$ 0.072 bits.

Streams S6, S4, and S3 form a high-importance core, with KL
divergences of 0.339, 0.306, and 0.278 respectively. Together, these
top streams account for approximately 50\% of the total causal
impact. A middle tier of streams (S7, S1, S5) contributes moderately,
while S0 and S2 exhibit the lowest importance.

Notably, all streams in the SRM-med model contribute meaningfully
(minimum KL $\approx$ 0.13), suggesting that at this scale the
architecture utilizes nearly all available stream capacity.

\begin{figure}[!htb]
  \centering
  \input{figs/161m_ablation.tex}
  \caption{Per-stream ablation impact for SRM-med, ranked by KL
    divergence. A small subset of streams accounts for a disproportionate
    share of causal contribution, while all streams exhibit non-zero
  impact.}
  \label{fig:161m_ablation}
\end{figure}
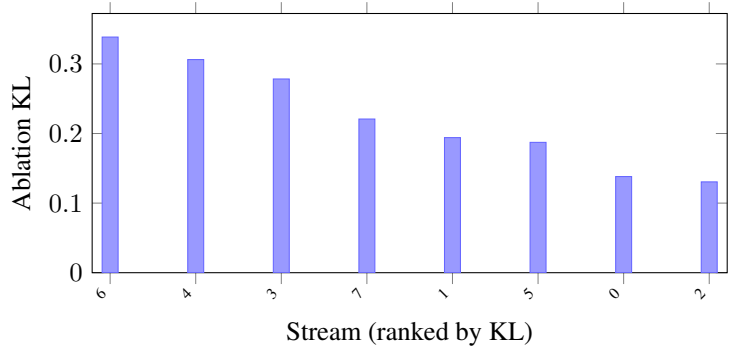

\subsection{Update Alignment}

\subsubsection{Attention-Token Alignment}

The SRM-med model exhibits a similar pattern to the baseline.
Attention-token cosine similarity remains near zero across layers
(mean 0.056, range -0.022 to 0.154), with a slight increase in later
layers (peak $\approx$ 0.074 at layer 5).

The persistence of near-orthogonal alignment at larger scale
indicates that attention-token decoupling is not an artifact of the
baseline configuration but a stable architectural property.

\begin{figure}[!htb]
  \centering
  \input{figs/161m_update_attn.tex}
  \caption{Cosine similarity between attention outputs and token inputs
    for SRM-med. Values cluster near zero across layers and streams,
    indicating that attention-derived updates are not aligned with token
  directions.}
  \label{fig:161m_update_attn}
\end{figure}
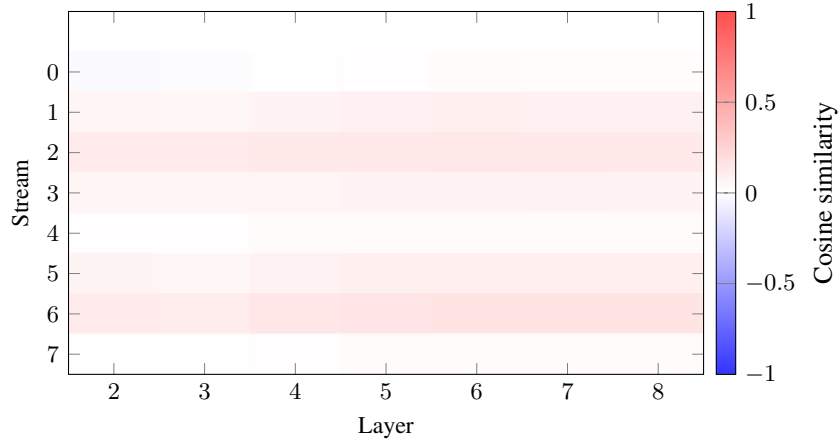

\subsubsection{Update-State Alignment}

In the SRM-med model, cosine similarity between the RMS-normalized
combined update and the RMS-normalized hidden stream state is
uniformly positive across all streams, layers, and recursive steps.
No systematic negative alignment is observed. Alignment values remain
high throughout the network.

Across recursive steps (ABC ordering), alignment exhibits clear
stepwise structure. Within each layer, alignment is stable and
remains strongly positive across steps, with modest increases at
later steps and deeper layers.

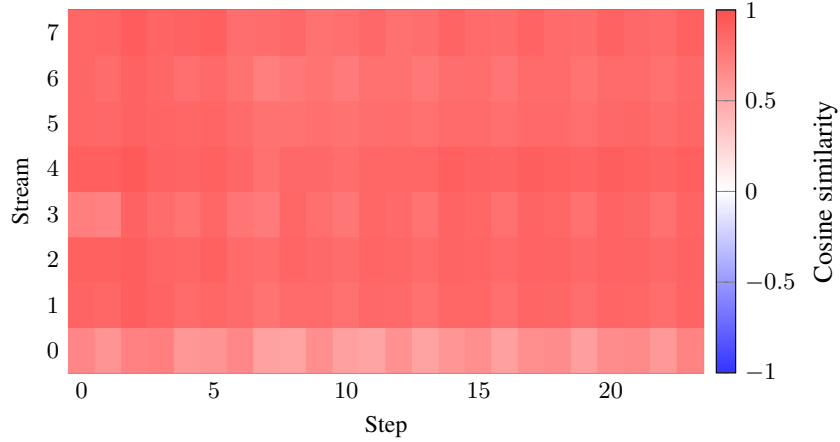
\begin{figure}[!htb]
  \centering
  \input{figs/161m_update_state.tex}
  \caption{Cosine similarity between the combined update and the
    existing stream state for SRM-med. Strong positive alignment across
    layers indicates recursive refinement rather than overwriting of
  stream representations.}
  \label{fig:161m_update_state}
\end{figure}

\subsection{Routing Structure}

The SRM-med model exhibits structured and non-uniform stream-to-stream
routing over a 10\% attention threshold. Routing activity is
concentrated in a subset of streams that consistently act as dominant
sources across layers and attention heads.

A prominent feature is strong diagonal dominance: within each head,
approximately 2-4 streams frequently route attention back to
themselves at rates exceeding 50\%, producing clear diagonal bands in
the routing matrices.

Routing structure varies systematically by attention head. Head H0
displays a hub-like pattern, with one or two streams routing broadly
to many targets. Head H2 exhibits more channelized behavior, where
routing is concentrated between specific stream pairs. Head H3 shows
more context-dependent routing, with less rigid structure and greater
dispersion across targets.

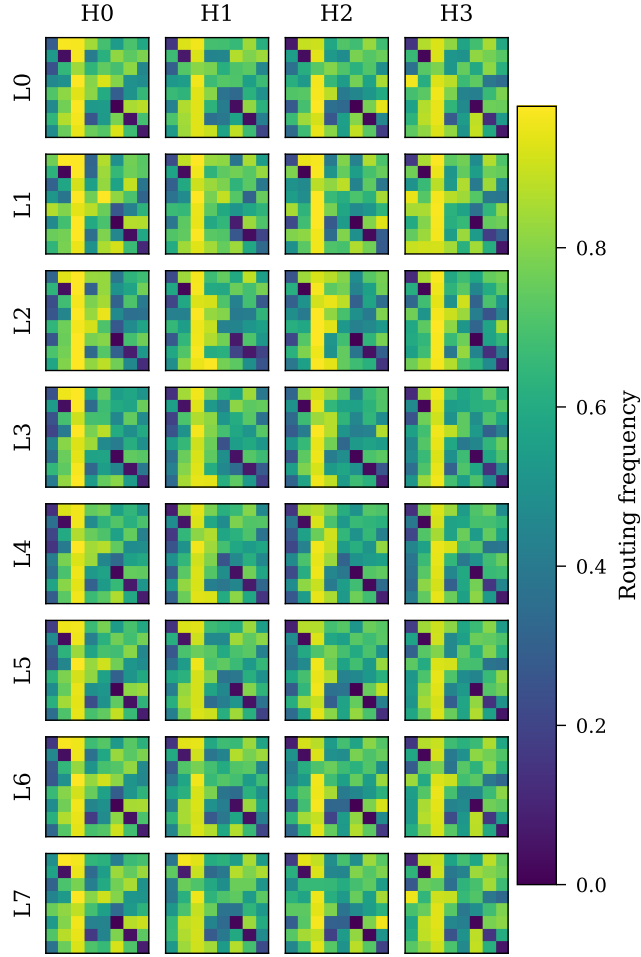
\begin{figure}[!htb]
  \centering
  \input{figs/161m_routing.tex}
  \caption{Stream-to-stream routing frequency for SRM-med at a 10\% attention threshold. Routing patterns are dense and structured across layers and heads, with evidence of hub-like streams and specialized routing structure.}
  \label{fig:161m_routing}
\end{figure}

\subsection{Stream-wise Logit Lens}

The SRM-med model exhibits clear and stable stratification in stream-level
decodability. While no individual stream is sufficient to recover the
model's prediction in isolation, a subset of streams becomes
progressively more aligned with the output space as depth increases.

Early layers (L0-L1) show uniformly weak decodability across all
streams. From L2 onward, streams begin to differentiate. Several
streams, most notably S6, S4, and S1, exhibit consistent and
monotonic improvement through mid and late layers, while others (S3,
S5, S7) remain weakly predictive throughout.

By the final layers (L6-L7), stratification is pronounced. Stream S6
achieves the strongest alignment with the output head (mean
log-probability $\approx$ -8.5), followed by S4 and S1. The relative
ordering of streams stabilizes in later layers.

\begin{figure}[!htb]
  \centering
  \input{figs/161m_logit_curves.tex}
  \caption{Per-stream logit lens curves for SRM-med showing the mean
    log-probability of the correct token across layers. Streams exhibit
    coordinated mid-depth degradation followed by partial recovery,
  with modest stream-to-stream variation.}
  \label{fig:161m_logit_curves}
\end{figure}
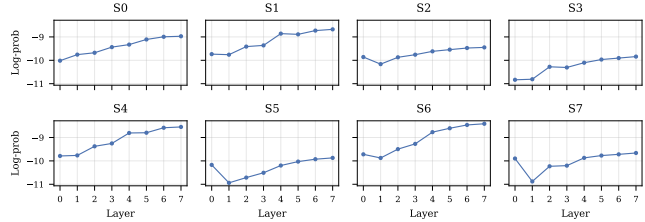

\begin{figure}[!htb]
  \centering
  \input{figs/161m_logit_heatmap.tex}
  \caption{Stream-wise logit lens heatmap for SRM-med showing the mean
    log-probability of the correct token across layers. A global mid-layer
  degradation is visible across nearly all streams.}
  \label{fig:161m_logit_heatmap}
\end{figure}
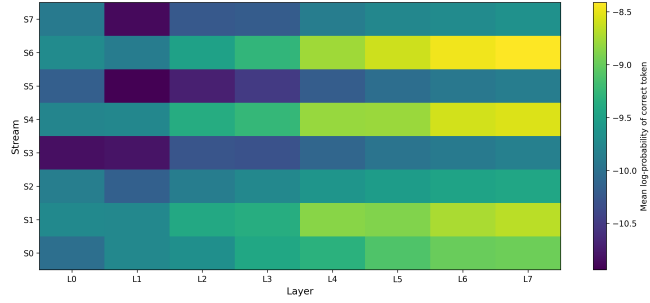

The heatmap representation (Figure~\ref{fig:161m_logit_heatmap})
provides a complementary view, showing the global mid-layer
degradation visible across nearly all streams.

\section{SRM-large Experiments and Results}
\label{sec:appendix_479m}

This appendix provides detailed experimental results for the SRM-large
model (16 streams, 512 dimensions each, ABCDEF step-group ordering).

\subsection{Representational Drift (Cosine Distance)}

As in the SRM-med model, KL divergence becomes numerically unstable for
512-dimensional stream representations in the SRM-large model. We use
cosine distance drift as the primary metric.

Across all six step groups (A-F), cosine drift values increase from
near zero at early transitions to moderate values at mid and late
depth, indicating sustained representational change across layers. In
contrast to the SRM-med model, no step group exhibits group-level early
stabilization across the majority of streams.

Convergence statistics reveal that late-converging streams dominate
in every step group. Group A shows mean convergence at 4.13 $\pm$
1.63 layers with 2 early, 1 middle, and 13 late convergers. Group B
converges at 4.25 $\pm$ 2.08 layers (3 early, 2 middle, 11 late).
Group C converges at 4.06 $\pm$ 2.11 layers (3 early, 3 middle, 10
late). Group D converges at 4.63 $\pm$ 2.09 layers (3 early, 1
middle, 12 late). Group E shows the earliest mean convergence at
3.44 $\pm$ 1.79 layers (2 early, 6 middle, 8 late). Group F
converges at 4.19 $\pm$ 1.64 layers (1 early, 4 middle, 11 late).

\begin{figure}[!htb]
  \centering
  \input{figs/479m_drift.tex}
  \caption{Per-stream representational drift measured by cosine distance across depth for the SRM-large model. Median trends and interquartile ranges reveal a pronounced mid-depth transition, consistent with coordinated representational change at scale.}
  \label{fig:479m_drift}
\end{figure}
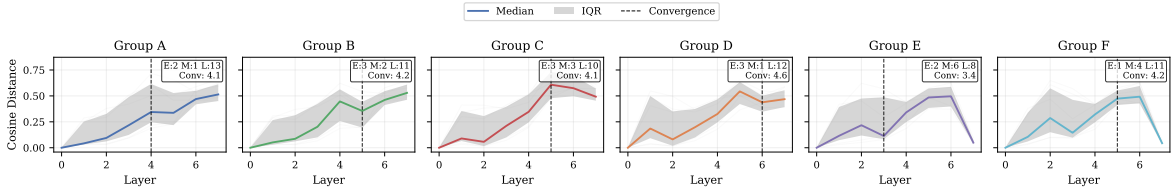

\subsection{Stream Ablation}

Mean ablation in the SRM-large model reveals hierarchical stratification
with a wider dynamic range than the SRM-med model. Ablation KL values
span 0.013 to 0.155 bits ($\approx$11.6$\times$).

Streams separate into four tiers based on ablation impact. The
high-importance tier comprises S5, S12, S3, and S6, which exhibit the
largest ablation effects (KL $\approx$ 0.12-0.16). The middle-high
tier includes S9, S11, S7, S1, and S15, producing moderate KL
increases ($\approx$ 0.07-0.10). The middle-low tier contains S8, S2,
S4, and S0, contributing modestly ($\approx$ 0.05-0.06). Finally,
S10, S13, and S14 form a minimal-impact tier, producing very small KL
changes ($<$0.03) and indicating near-redundant contributions.

Mean ablation KL per stream drops approximately 65\% from SRM-med to
SRM-large (0.224 to 0.078 bits), indicating that causal influence
spreads across more streams even as relative differentiation increases.

\begin{figure}[!htb]
  \centering
  \input{figs/479m_ablation.tex}
  \caption{Per-stream ablation impact for SRM-large, ranked by KL
    divergence. A highly skewed distribution indicates that a small
  subset of streams contributes a large fraction of causal influence.}
  \label{fig:479m_ablation}
\end{figure}
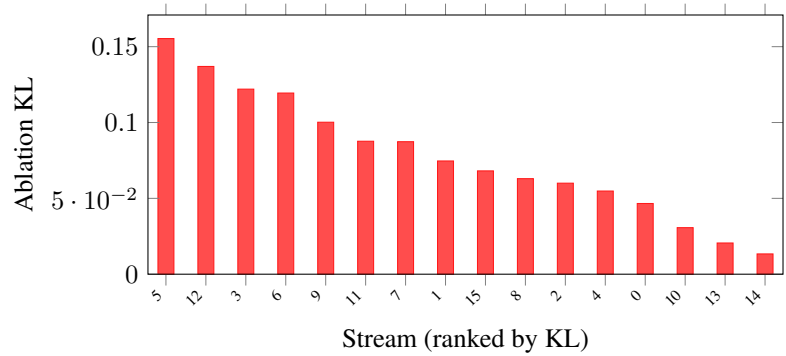

\subsection{Update Alignment}

\subsubsection{Attention-Token Alignment}

The SRM-large model maintains near-zero attention-token alignment despite
increased stream count and deeper recursion. Mean cosine similarity
is 0.064 (range -0.015 to 0.171), with a layer-wise peak at layer 5
($\approx$ 0.082).

The close correspondence between SRM-med and SRM-large values demonstrates
that attention-token orthogonality is invariant to scale and stream count.

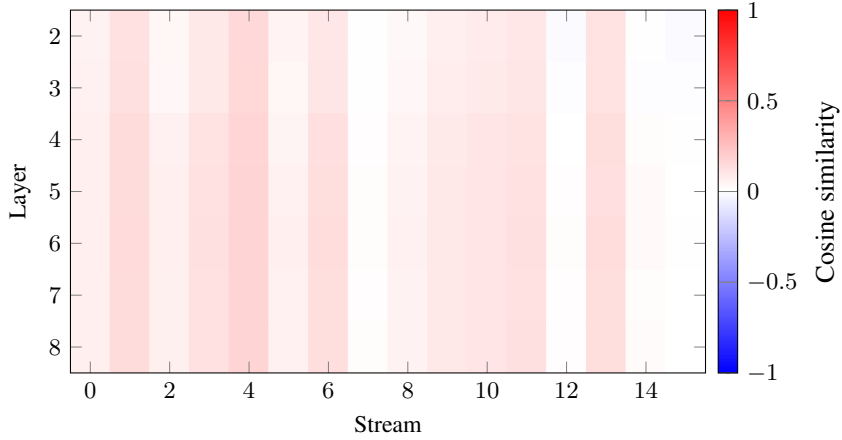
\begin{figure}[!htb]
  \centering
  \input{figs/479m_update_attn.tex}
  \caption{Cosine similarity between attention outputs and token inputs
    for SRM-large. Near-zero alignment across layers and streams indicates
  that attention introduces directions distinct from token embeddings.}
  \label{fig:479m_update_attn}
\end{figure}

\subsubsection{Update-State Alignment}

In the SRM-large model, update-state alignment remains predominantly
positive across streams and layers. Most streams exhibit strong
alignment between the combined update and the existing hidden state.

However, unlike the baseline and SRM-med model, a distinct band of
reduced or negative alignment is consistently observed for a small
subset of streams across specific recursive steps. This appears as a
pronounced stripe in the heatmap, indicating updates that are
directionally opposed to the current stream state. The persistence of
this pattern across depth suggests systematic behavior rather than noise.

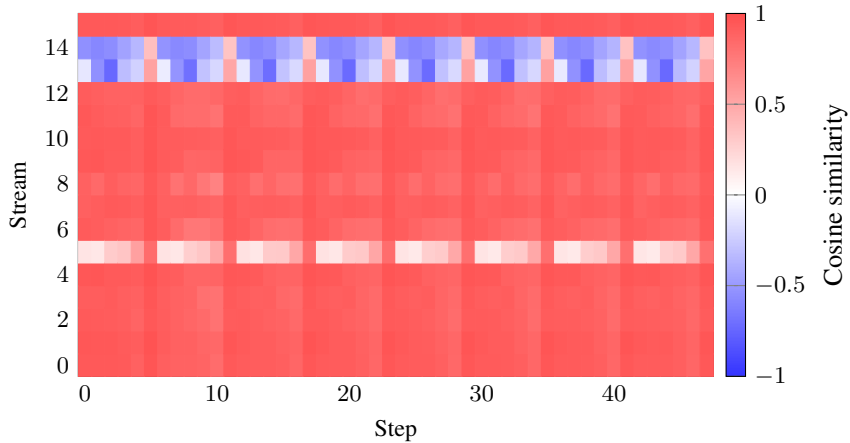
\begin{figure}[!htb]
  \centering
  \input{figs/479m_update_state.tex}
  \caption{Cosine similarity between the combined update and existing
    stream state for SRM-large. Strong positive alignment indicates
  recursive refinement is preserved at larger scale.}
  \label{fig:479m_update_state}
\end{figure}

\subsection{Routing Structure}

The SRM-large model exhibits sparse and highly selective stream-to-stream
routing. In contrast to the denser routing in the SRM-med model, routing
events above threshold are concentrated in a small number of stream pairs.

Self-routing is reduced relative to the SRM-med model. Within each head,
only 1-3 streams consistently exhibit elevated self-attention,
producing weaker diagonal structure.

Routing structure is characterized by focal high-frequency edges.
Rather than broad bands of elevated routing, the matrices show
isolated, high-intensity entries corresponding to specific
source-target stream pairs. These focal connections persist across layers.

The increase in attention heads (8 versus 4) corresponds to greater
diversification of routing patterns. Individual heads specialize over
narrower subsets of stream pairs.

\begin{figure}[!htb]
  \centering
  \includegraphics[width=0.9\columnwidth]{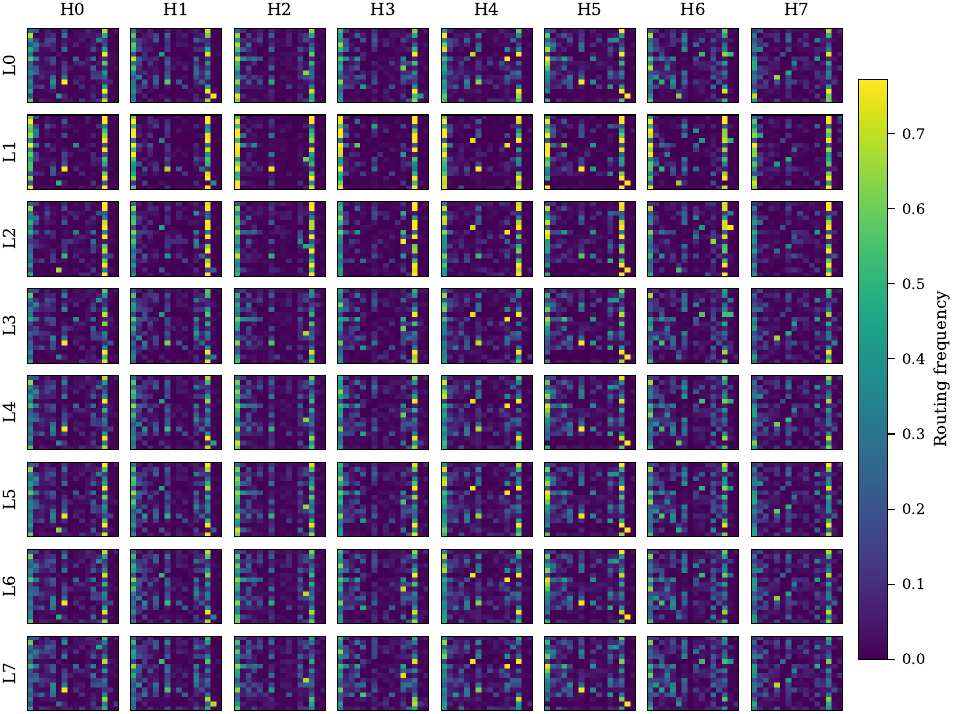}
  \caption{Stream-to-stream routing frequency for SRM-large at a 10\% attention threshold. Routing is sparse and highly selective across layers and heads, with a small number of dedicated, high-frequency stream-to-stream channels.}
  \label{fig:479m_routing}
\end{figure}

\subsection{Stream-wise Logit Lens}

Under mean ablation, the SRM-large model exhibits strong and rapidly
emerging stream-level decodability. Unlike smaller models, alignment
with the output space develops early and remains stable across depth.

Early layers show uniformly weak decodability at L0, followed by a
sharp improvement at L1 across nearly all streams. From L2 onward,
mean log-probabilities cluster tightly, with only modest variation
between streams. This narrow performance band persists through mid
and late layers.

No stream exhibits sustained degradation or failure. All streams
follow similar trajectories: a steep initial gain, followed by
gradual refinement and stabilization. The overall ordering does not
change substantially with depth.

\begin{figure}[!htb]
  \centering
  \input{figs/479m_logit_curves.tex}
  \caption{Per-stream logit lens curves for SRM-large showing the mean
    log-probability of the correct token across layers. Despite increased
    model scale, streams exhibit coordinated degradation and partial
  recovery patterns.}
  \label{fig:479m_logit_curves}
\end{figure}
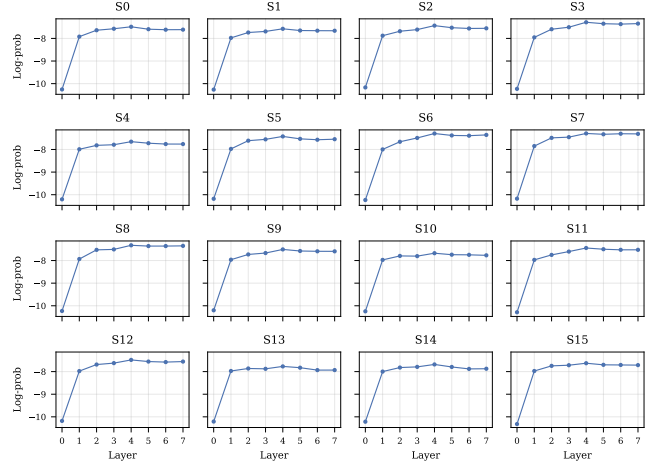

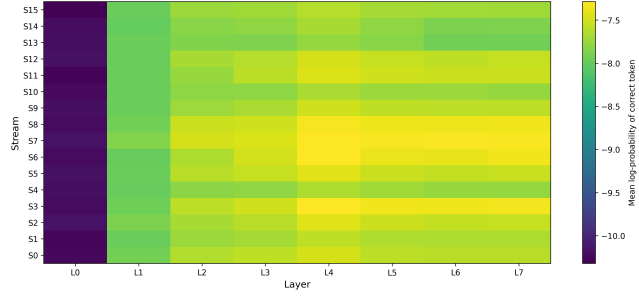
\begin{figure}[!htb]
  \centering
  \input{figs/479m_logit_heatmap.tex}
  \caption{Stream-wise logit lens heatmap for SRM-large showing the mean
    log-probability of the correct token across layers. A pronounced
  mid-layer degradation is visible across nearly all streams.}
  \label{fig:479m_logit_heatmap}
\end{figure}

The heatmap representation (Figure~\ref{fig:479m_logit_heatmap})
confirms the pronounced mid-layer degradation visible across nearly all streams.

%% file: figs/baseline_kl_three_panel.tex
\centering
\includegraphics[width=0.85\columnwidth, height=0.22\columnwidth,]{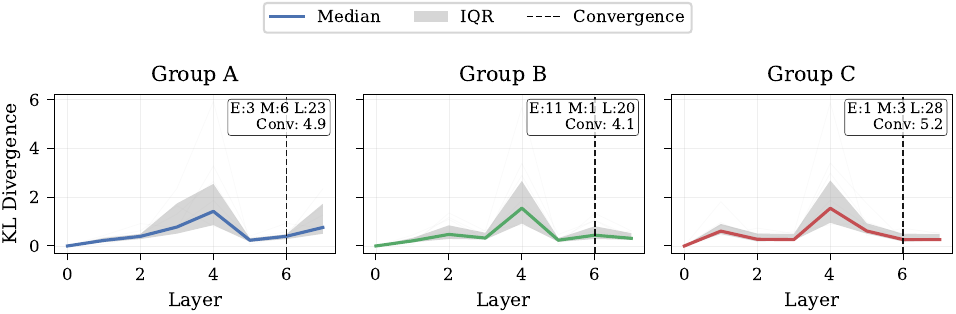}

%% file: figs/baseline_ablation.tex
\centering
\begin{tikzpicture}
  \begin{axis}[
    ybar,
    scale only axis,
    width=0.49\columnwidth,
    height=0.20\columnwidth,
    xlabel={Stream (ranked by KL)},
    ylabel={Ablation KL},
    ymin=0,
    xtick=data,
    xticklabels from table={data/baseline_ablation_ranked.dat}{stream},
    xticklabel style={rotate=45, anchor=east, font=\tiny},
    bar width=0.6em,
    enlarge x limits=0.03,
    axis on top,
    clip=true
  ]
  \addplot[fill=blue!40, draw=blue!60] table[x expr=\coordindex, y=kl] {data/baseline_ablation_ranked.dat};
  \end{axis}
\end{tikzpicture}

%% file: figs/baseline_update_attn.tex
\centering
\begin{tikzpicture}
  \begin{axis}[
    view={0}{90},
    scale only axis,
    width=0.32\columnwidth,
    height=0.28\columnwidth,
    colormap={rb}{color(0)=(blue) color(1)=(white) color(2)=(red)},
    point meta min=-1,
    point meta max=1,
    xlabel={Stream},
    ylabel={Layer},
    xmin=-0.5, xmax=31.5,
    ymin=1.5, ymax=8.5,
    ytick={1,2,3,4,5,6,7,8},
    tick label style={font=\small},
    label style={font=\small},
    colorbar,
    colorbar style={
      ylabel={Cosine similarity},
      width=0.015\textwidth,
      at={(1.02,0.5)},
      anchor=west
    },
    axis on top,
    clip=true
  ]
  \addplot[matrix plot, mesh/cols=32, point meta=explicit]
    table[x=stream, y=layer, meta=value] {data/baseline_attn_token_cos.dat};
  \end{axis}
\end{tikzpicture}

%% file: figs/baseline_update_state.tex
\centering
\begin{tikzpicture}
  \begin{axis}[
    view={0}{90},
    scale only axis,
    width=0.32\columnwidth,
    height=0.28\columnwidth,
    colormap={paper}{color(0)=(blue!80) color(0.5)=(white) color(1)=(red!70)},
    point meta min=-1,
    point meta max=1,
    xlabel={Stream},
    ylabel={Step},
    xmin=-0.5, xmax=31.5,
    ymin=0, ymax=47.5,
    tick label style={font=\small},
    label style={font=\small},
    colorbar,
    colorbar style={
      ylabel={Cosine similarity},
      width=0.015\textwidth,
      at={(1.02,0.5)},
      anchor=west
    },
    axis on top,
    clip=true
  ]
  \addplot[matrix plot, mesh/cols=32, point meta=explicit]
    table[x=stream, y=step, meta=value] {data/baseline_update_state_cos.dat};
  \end{axis}
\end{tikzpicture}

%% file: figs/baseline_routing.tex
\centering
\includegraphics[width=0.49\columnwidth]{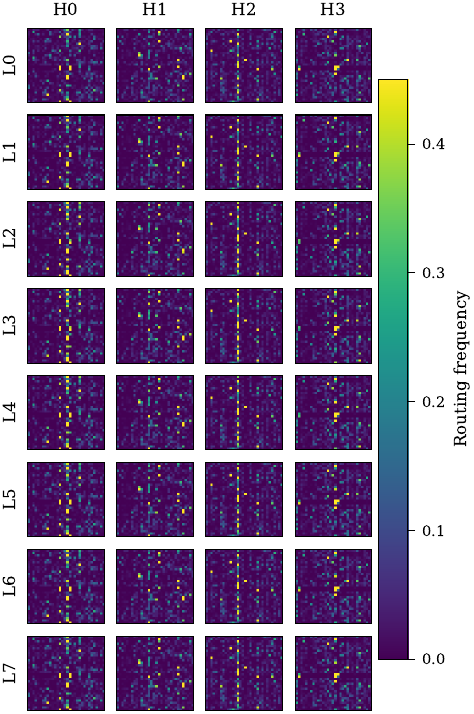}

%% file: figs/baseline_logit_curves.tex
\centering
\includegraphics[width=0.49\columnwidth]{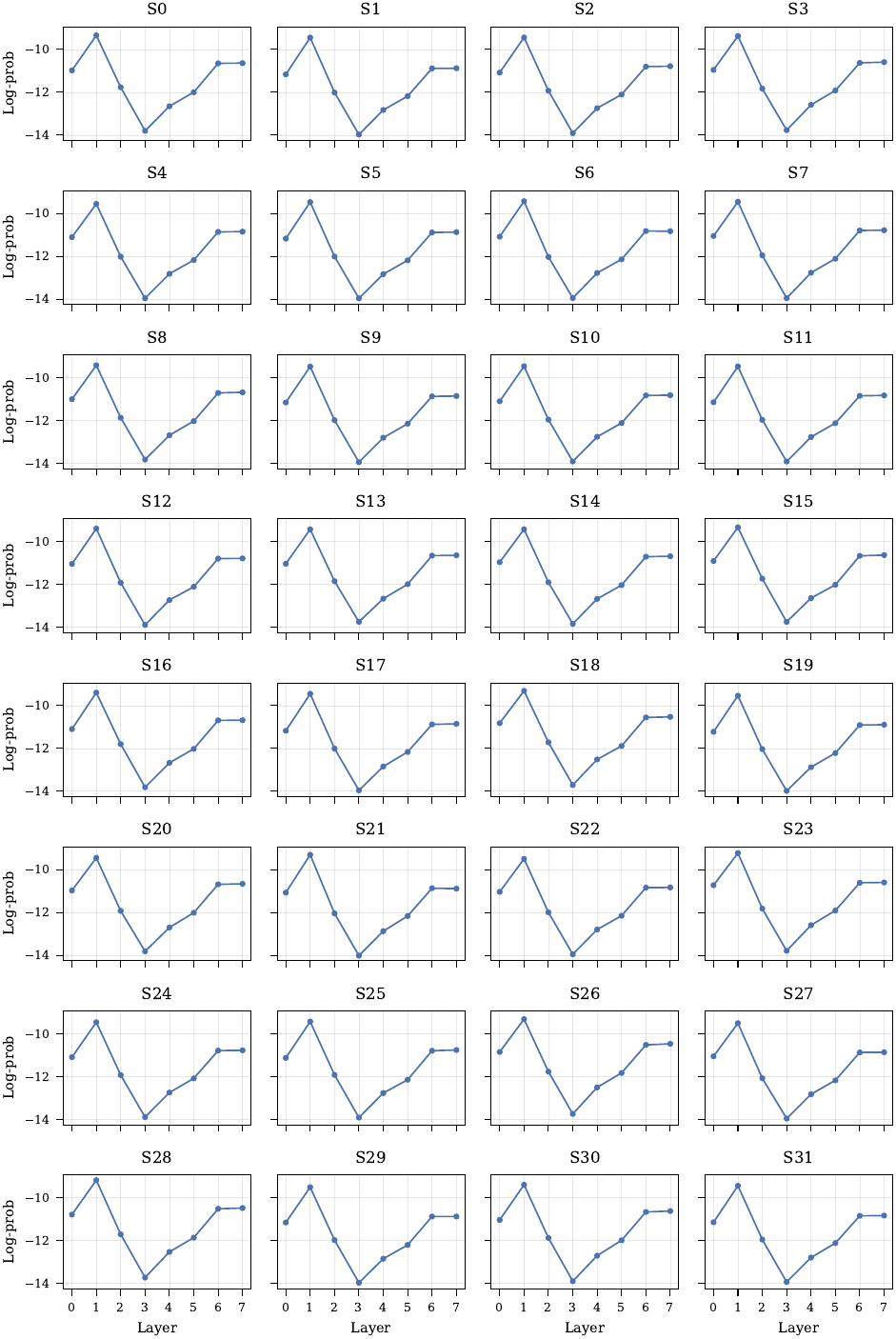}

%% file: figs/baseline_logit_heatmap.tex
\centering
\includegraphics[width=0.49\columnwidth, trim=0pt 0pt 0pt 35pt, clip]{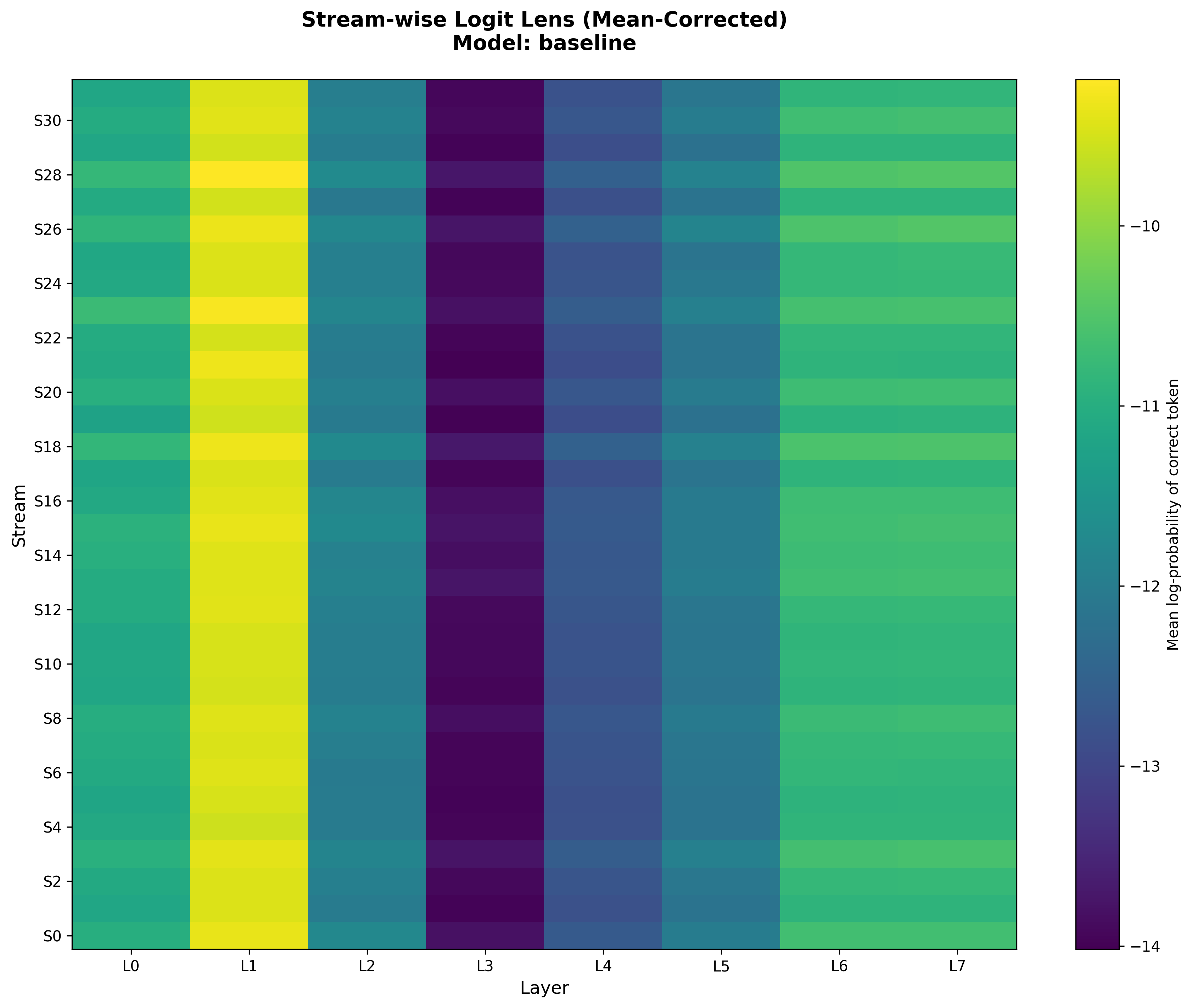}

%% file: figs/161m_drift.tex
\centering
\includegraphics[width=0.85\columnwidth]{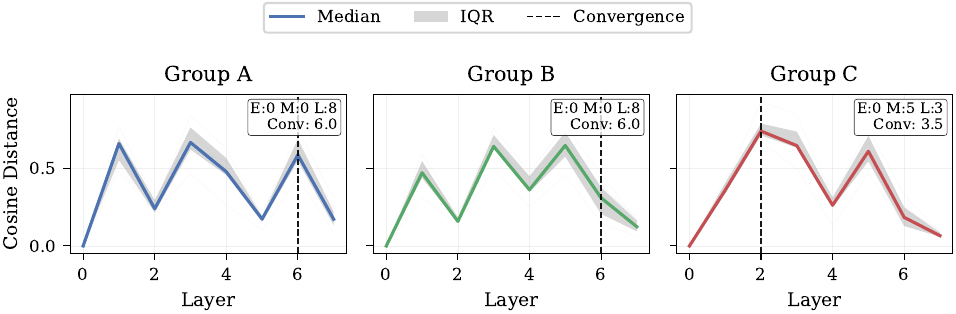}

%% file: figs/161m_ablation.tex
\centering
\begin{tikzpicture}
  \begin{axis}[
    ybar,
    scale only axis,
    width=0.49\columnwidth,
    height=0.20\columnwidth,
    xlabel={Stream (ranked by KL)},
    ylabel={Ablation KL},
    ymin=0,
    xtick=data,
    xticklabels from table={data/161m_ablation_ranked.dat}{stream},
    xticklabel style={rotate=45, anchor=east, font=\tiny},
    bar width=0.6em,
    enlarge x limits=0.03
  ]
  \addplot[fill=blue!40, draw=blue!60] table[x expr=\coordindex, y=kl] {data/161m_ablation_ranked.dat};
  \end{axis}
\end{tikzpicture}

%% file: figs/161m_update_attn.tex
\centering
\begin{tikzpicture}
  \begin{axis}[
      view={0}{90},
      scale only axis,
      width=0.49\columnwidth,
      height=0.28\columnwidth,
      colormap={paper}{color(0)=(blue!80) color(0.5)=(white) color(1)=(red!70)},
      point meta min=-1,
      point meta max=1,
      xlabel={Layer},
      ylabel={Stream},
      xmin=1.5, xmax=8.5,
      ymin=-1.5, ymax=7.5,
      xtick={1,2,3,4,5,6,7,8},
      ytick={0,1,2,3,4,5,6,7},
      tick label style={font=\small},
      label style={font=\small},
      colorbar,
      colorbar style={
        ylabel={Cosine similarity},
        width=0.015\textwidth,
        at={(1.02,0.5)},
        anchor=west
      }
    ]
    \addplot[matrix plot, mesh/cols=8, point meta=explicit]
    table[x=layer, y=stream, meta=value] {data/161m_attn_token_cos.dat};
  \end{axis}
\end{tikzpicture}

%% file: figs/161m_update_state.tex
\centering
\begin{tikzpicture}
  \begin{axis}[
      view={0}{90},
      scale only axis,
      width=0.49\columnwidth,
      height=0.28\columnwidth,
      colormap={paper}{color(0)=(blue!80) color(0.5)=(white) color(1)=(red!70)},
      point meta min=-1,
      point meta max=1,
      xlabel={Step},
      ylabel={Stream},
      xmin=-0.5, xmax=23.5,
      ymin=-0.5, ymax=7.5,
      ytick={0,1,2,3,4,5,6,7},
      tick label style={font=\small},
      label style={font=\small},
      colorbar,
      colorbar style={
        ylabel={Cosine similarity},
        width=0.015\textwidth,
        at={(1.02,0.5)},
        anchor=west
      }
    ]
    \addplot[matrix plot*, mesh/cols=8, point meta=explicit]
    table[x=step, y=stream, meta=value] {data/161m_update_state_cos.dat};
  \end{axis}
\end{tikzpicture}

%% file: figs/161m_routing.tex
\centering
\includegraphics[width=0.49\columnwidth]{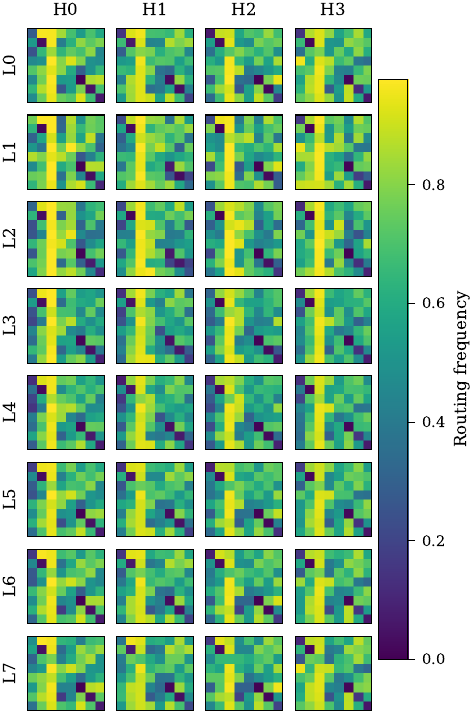}

%% file: figs/161m_logit_curves.tex
\centering
\includegraphics[width=0.49\columnwidth]{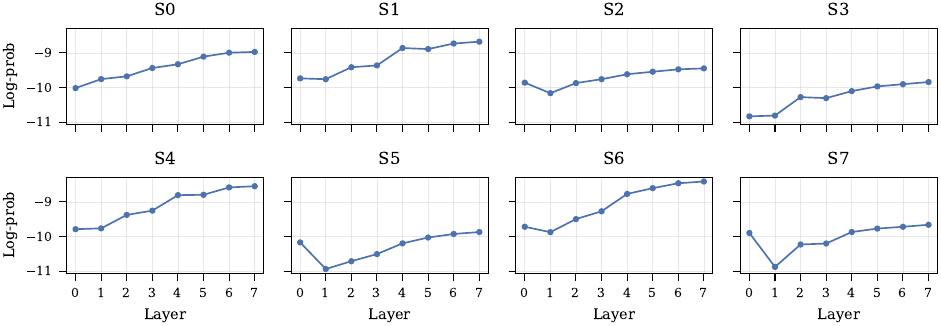}

%% file: figs/161m_logit_heatmap.tex
\centering
\includegraphics[width=0.5\columnwidth, trim=0pt 0pt 0pt 35pt,
clip]{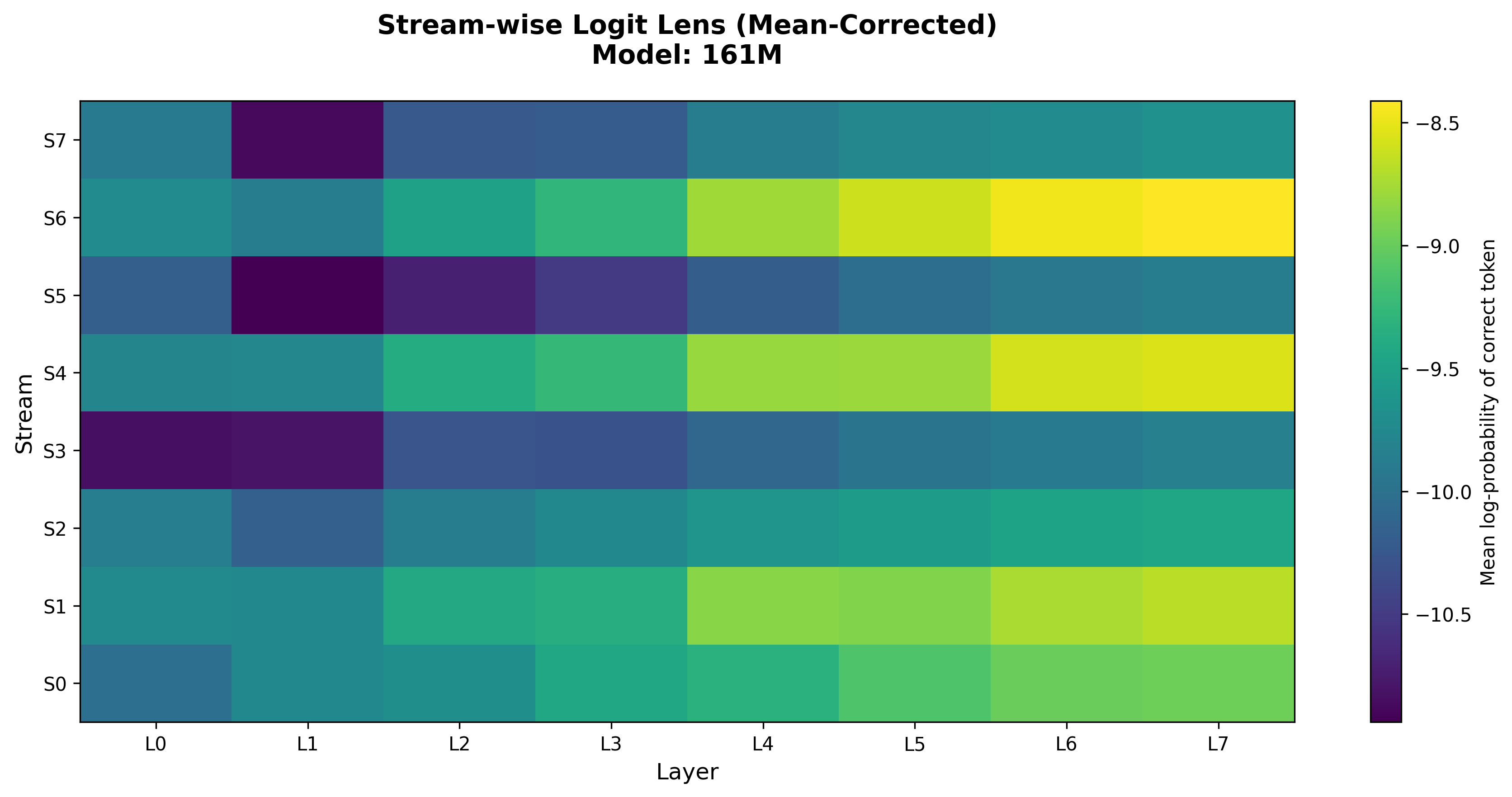}

%% file: figs/479m_drift.tex
\centering
\includegraphics[width=0.9\columnwidth]{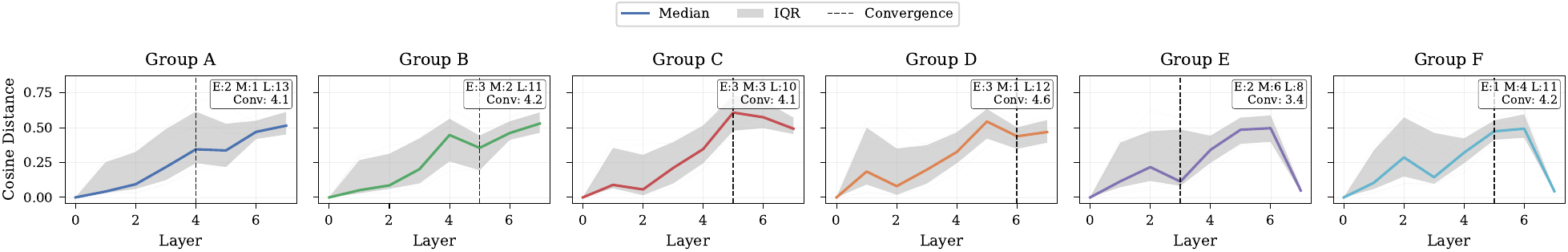}

%% file: figs/479m_ablation.tex
\centering
\begin{tikzpicture}
  \begin{axis}[
    ybar,
    scale only axis,
    width=0.49\columnwidth,
    height=0.20\columnwidth,
    xlabel={Stream (ranked by KL)},
    ylabel={Ablation KL},
    ymin=0,
    xtick=data,
    xticklabels from table={data/479m_ablation_ranked.dat}{stream},
    xticklabel style={rotate=45, anchor=east, font=\tiny},
    bar width=0.6em,
    enlarge x limits=0.03,
    axis on top,
    clip=true
  ]
  \addplot[fill=red!70, draw=red!90] table[x expr=\coordindex, y=kl] {data/479m_ablation_ranked.dat};
  \end{axis}
\end{tikzpicture}

%% file: figs/479m_update_attn.tex
\centering
\begin{tikzpicture}
  \begin{axis}[
    view={0}{90},
    scale only axis,
    width=0.49\columnwidth,
    height=0.28\columnwidth,
    colormap={rb}{color(0)=(blue) color(1)=(white) color(2)=(red)},
    point meta min=-1,
    point meta max=1,
    xlabel={Stream},
    ylabel={Layer},
    xmin=-0.5, xmax=15.5,
    ymin=1.5, ymax=8.5,
    ytick={1,2,3,4,5,6,7,8},
    tick label style={font=\small},
    label style={font=\small},
    colorbar,
    colorbar style={
      ylabel={Cosine similarity},
      width=0.015\textwidth,
      at={(1.02,0.5)},
      anchor=west
    }
  ]
  \addplot[matrix plot, mesh/cols=16, point meta=explicit]
    table[x=stream, y=layer, meta=value] {data/479m_attn_token_cos.dat};
  \end{axis}
\end{tikzpicture}

%% file: figs/479m_update_state.tex
\centering
\begin{tikzpicture}
  \begin{axis}[
    view={0}{90},
    scale only axis,
    width=0.49\columnwidth,
    height=0.28\columnwidth,
    colormap={paper}{color(0)=(blue!80) color(0.5)=(white) color(1)=(red!70)},
    point meta min=-1,
    point meta max=1,
    xlabel={Step},
    ylabel={Stream},
    xmin=-0.5, xmax=47.5,
    ymin=-0.5, ymax=15.5,
    ytick={0,2,4,6,8,10,12,14},
    tick label style={font=\small},
    label style={font=\small},
    colorbar,
    colorbar style={
      ylabel={Cosine similarity},
      width=0.015\textwidth,
      at={(1.02,0.5)},
      anchor=west
    }
  ]
  \addplot[matrix plot*, mesh/cols=16, point meta=explicit]
    table[x=step, y=stream, meta=value] {data/479m_update_state_cos.dat};
  \end{axis}
\end{tikzpicture}

%% file: figs/479m_logit_curves.tex
\centering
\includegraphics[width=0.49\columnwidth]{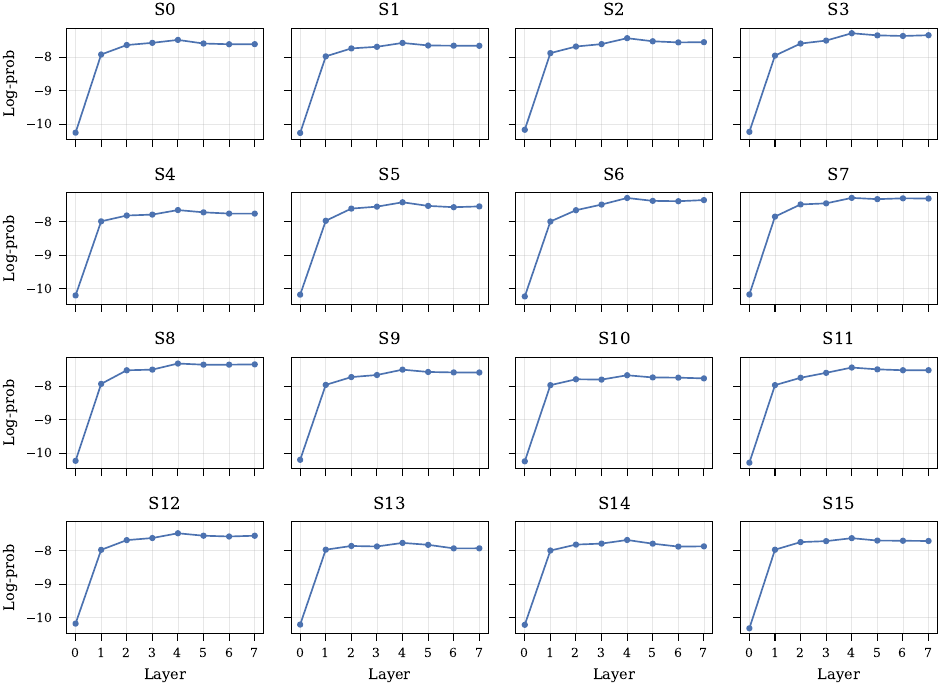}

%% file: figs/479m_logit_heatmap.tex
\centering
\includegraphics[width=0.49\columnwidth, trim=0pt 0pt 0pt 35pt,
clip]{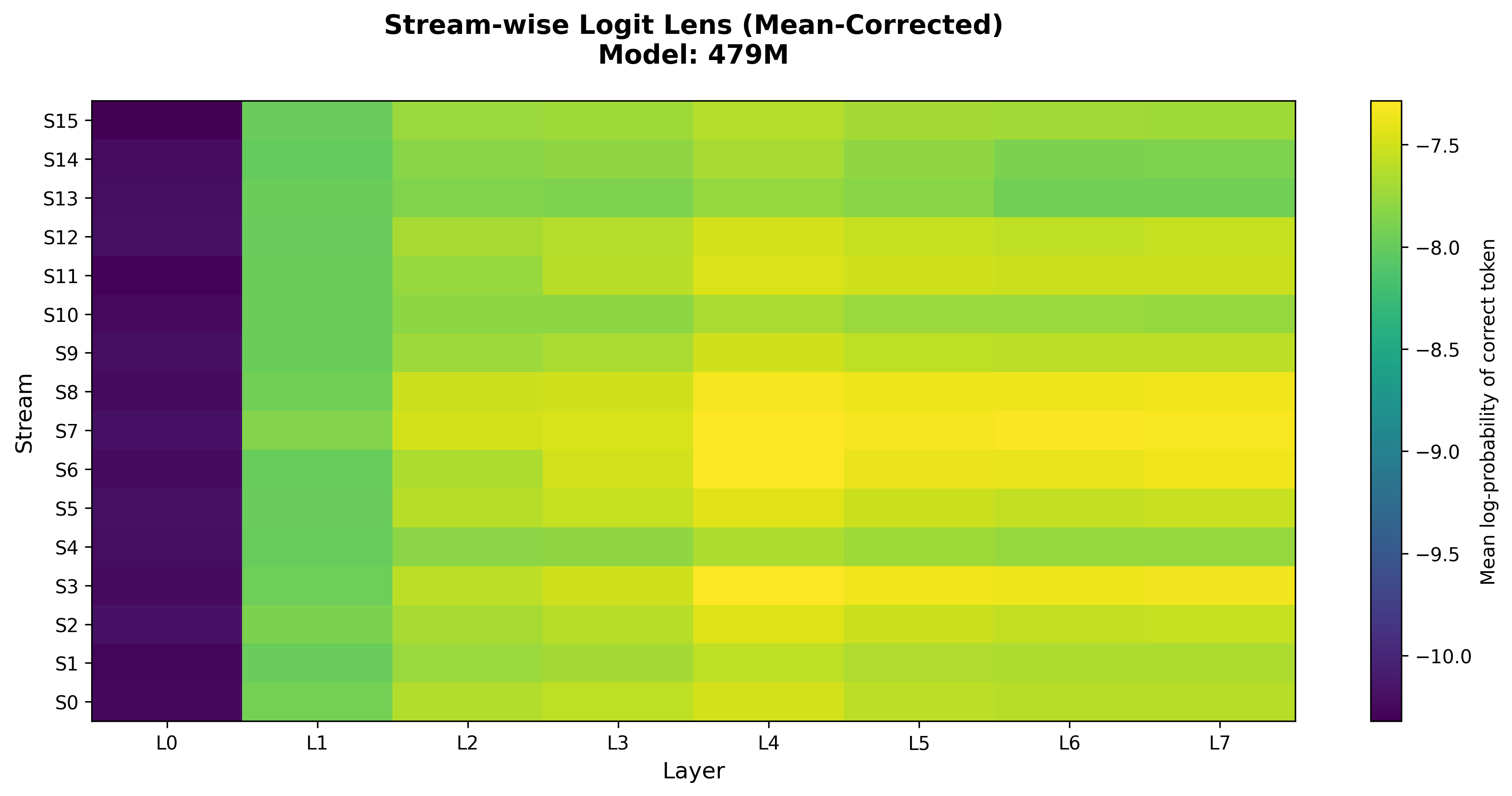}